\documentclass[preprint,12pt,authoryear]{elsarticle}
\makeatletter 
\def\ps@pprintTitle{} 
\makeatother

\usepackage{amssymb}
\usepackage{amsmath}

\usepackage{amsfonts}
\usepackage{array}
\usepackage{textcomp}
\usepackage{stfloats}
\usepackage{url}
\usepackage{verbatim}
\usepackage{graphicx}

\usepackage{caption}
\usepackage{subcaption}
\usepackage{booktabs}
\usepackage{multirow}
\newtheorem{theorem}{Theorem}
\usepackage{algorithmic}
\usepackage{algorithm}
\usepackage{xcolor}
\usepackage{hyperref}
\newtheorem{remark}{Remark}

\journal{International Journal of Forecasting}

\begin{document}

\begin{frontmatter}



\title{Deep Switching State Space Model for Nonlinear Time Series Forecasting with Regime Switching} 


\author[label1]{Xiuqin Xu, Hanqiu Peng, and Ying Chen}
\affiliation[label1]{
	organization={Department of Mathematics, Asian Institute of Digital Finance, and Risk Management Institute},
	addressline={National University of Singapore}, 
	city={21 Lower Kent Ridge Rd},
	postcode={119077}, 
country={Singapore}
} 

\newpage

\begin{abstract}
Modern time series data often display complex nonlinear dependencies along with irregular regime-switching behaviors. These features present technical challenges in modeling, inference, and providing insightful understanding of the underlying stochastic phenomena. 
To tackle these challenges, we introduce the novel Deep Switching State Space Model (DS$^3$M). 
In DS$^3$M, the architecture employs discrete latent variables to represent regimes and continuous latent variables to account for random driving factors. 
By melding a Recurrent Neural Network (RNN) with a nonlinear Switching State Space Model (SSSM), we manage to capture the nonlinear dependencies and irregular regime-switching behaviors, governed by a Markov chain and parameterized using multilayer perceptrons. 
We validate the DS$^3$M through short- and long-term forecasting on a wide array of simulated and real-world datasets, spanning sectors such as healthcare, economics, traffic, meteorology, and energy. 
Our results reveal that DS$^3$M outperforms several state-of-the-art models in terms of forecasting accuracy, while providing meaningful regime identifications.

\end{abstract}


%


\end{frontmatter}



\section{Introduction}
\label{sec:introduction}

In many studies, researchers face challenges in modeling and inferring from modern time-series data collected across various disciplines. Examples include healthcare (such as sleep apnea), economics (unemployment rates), traffic and transportation (metro passenger volume), meteorology (sea surface temperature), and energy (electricity demand), among others. In these contexts, the conventional assumptions of stationarity and linearity that often form the foundation of statistical modeling are frequently inadequate.  Instead, they are replaced by complex nonlinear dynamics intertwined with irregular regime shifts \citep{segnon2024forecasting}. For instance, the unemployment rate is influenced by economic conditions like booms or recessions. It is also affected by latent continuous variables, such as regional wage elasticity, which vary with the discrete regime states.  The transition of the latent variables between different steps and the interactions between the latent variables and the observations could also be nonlinear. In these settings, both conventional linear statistical models and standard deep learning approaches suffer from either severe modeling misspecification or a lack of effective identification of meaningful stochastic regimes.

Given the presence of regimes within such time series data, Switching State Space Models (switching SSMs) are arguably the most widely used. In these models, the evolution of the time series is presumed to be driven by hidden factors that switch among discrete regimes \citep{ghahramani2000variational,fox2009nonparametric}.  
The dynamics within each regime are typically represented by simple linear models that can be efficiently estimated even with a limited sample size \citep{durbin2012time}, while the transitions between regimes are governed by the hidden transition probabilities of a Markov chain. For complex dependence where the stage-wise linear structure is insufficient, SSMs can be customized with certain pre-specified nonlinear transition and/or emission functions \citep{chow2013nonlinear}. 
Despite their popularity, existing nonlinear models rely on predetermined local parametric forms with simple structures, either linear or nonlinear, which may be insufficient to describe the actual patterns in modern nonlinear time series, leading to the misspecification problem.

On the contrary, deep learning methods, especially recurrent neural networks \citep{hewamalage2021recurrent} with gate structures, such as the Long-Short Term Memory (LSTM, \citep{hochreiter1997long}), Gated Recurrent Unit (GRU, \citep{chung2014empirical}), Transformers \citep{ chen2024contiformer, liu2024itransformer}, and temporal convolution networks \citep{sen2019think}, have emerged as the new benchmark to model nonlinear time series with highly complex dependencies. However, the classic deep learning models are deterministic and ignore the presence of unobserved stochastic signals in the dynamic system. The only randomness allowed appears in the conditional output probability models. As deterministic models often struggle to capture stochastic behavior in nonlinear time series with non-stationary patterns and transitions \citep{zhang2005neural}, they require a large number of parameters to ensure consistent estimation and accurate modeling. However, this is typically impractical due to the limited availability of real-world time series data in many disciplines. Additionally, deterministic deep learning models lack interpretability when applied to stochastic dynamic systems.

This has led to the integration of deep learning and stochastic latent variable models to leverage their complementary strengths of nonlinear representation and interpretability \citep{kingma2013auto, rezende2014stochastic, bayer2014learning}. 
Deep State Space Models (SSMs) often introduce continuous Gaussian latent variables at each time step to capture latent dynamics, essentially functioning as sequences of variational auto-encoders \citep{krishnan2017structured, fraccaro2017disentangled, de2020normalizing, girin2020dynamical}. 
While effective for prediction, these models struggle with interpretability in non-stationary time series due to the absence of discrete regimes.  
Some deep SSMs introduce discrete latent variables to model regime switching \citep{johnson2016composing, dai2016recurrent, Liu2018}.  Yet, these approaches often oversimplify by assuming that time series evolution is solely driven by discrete variables, neglecting the interplay between discrete and continuous latent factors, which is crucial for capturing real-world dynamics.

We propose a Deep Switching State Space Model (DS$^3$M)for interpretable and efficient inference in nonlinear time series with irregular regime switching. 
DS$^3$M is designed to provide accurate forecasts while identifying hidden regimes with significant economic implications. 
The architecture combines a Recurrent Neural Network (RNN) with a Nonlinear Switching State Space Model (SSSM), where regime-switching is governed by a Markov chain.  The model uses discrete latent variables to represent regimes and continuous latent variables for random driving factors, capturing the joint impact of these on the time series dynamics. Discrete latent variables influence both the observed data and continuous latent variables.
The RNN enhances forecasting accuracy by leveraging long-term information and skip connections to observations. We develop an approximate variational inference algorithm that is scalable to large datasets. The key idea involves marginalizing the discrete latent variables solely at each time step and subsequently utilizing a reparametrization trick for the continuous latent variables. Applied to various domains—healthcare, economics, traffic, meteorology, and energy—DS$^3$M outperforms state-of-the-art methods (e.g., GRU, SRNN \citep{fraccaro2016sequential}, DSARF \citep{farnoosh2020deep}, SNLDS \citep{dong2020collapsed}, offering better predictive accuracy and meaningful regime identification. Notably, DS$^3$M identifies longer regime durations, aligning more closely with real-world data compared to the chaotic switching seen in other models.

Our paper contributes a recipe for employing statistical modeling and deep learning to achieve interpretable inference for modern time series with complex dynamics. By introducing both continuous and discrete latent variables in the recurrent neural network, we efficiently harness the potent representation capabilities of continuous latent variables to capture rich dependence and enable economically meaningful identifications of discrete latent variables. The novel amortized variational inference method renders it suitable for both small and large datasets. When applied to a variety of simulated and real data across various disciplines, we showcase the robust competitive performance of DS$^3$M compared to state-of-the-art technologies.

One of the most relevant models is the SNLDS, proposed by \cite{dong2020collapsed}. The key difference between SNLDS and our DS$^3$M model lies in the design of both the generative model and inference algorithm. In our approach, we integrate a recurrent neural network (RNN) into DS$^3$M to capture historical information and introduce dependencies between the observations, continuous latent variables, and the hidden states of the RNN. This enables us to address non-Markov problems within a Markov framework. The design allows the RNN and continuous latent variables to encode different types of information, leading to improved prediction accuracy, as demonstrated in our experiments. Additionally, we developed a more stable inference algorithm to mitigate posterior collapse—a situation where discrete latent variables are underutilized because the probabilistic model for continuous latent variables is overly powerful. Unlike \cite{dong2020collapsed}, where the approximate posterior of the continuous latent variables is independent of the discrete latent variables, our approach employs connected inference structures that better mimic the true posterior relationships.

The rest of the paper is organized as follows.
Section \ref{sec:background} reviews the related works. 
Section \ref{sec:formulation} details the proposed DS$^3$M and the scalable inference algorithm. 
Section \ref{sec:simulation} presents the numerical performance of the DS$^3$M with several simulated and six real-world data sets in different disciplines. 
Section \ref{sec:conclusion} concludes. 
Data and codes can be found on the \href{https://github.com/Sherry-Xu/Deep-Switching-State-Space-Model}{Github website}.

\section{Background and related works}
\label{sec:background} 

Denote a time series of $T$ observations as $y_{1:T}=\{y_1,y_2,\cdots,y_T\}$, $y_t \in \mathbb{R}^D$ and a sequence of inputs as $x_{1:T} = \{x_1,x_2,\cdots,x_T\}$, $x_t \in \mathbb{R}^U$. 
In the setting of time series forecasting, $x_t$  can be one or multiple lagged values of the time series, e.g. $y_{t-1}$ and higher orders $y_{t-2}, y_{t-3}, \cdots$.
The inputs $x_t$ could also contain exogenous variables. 
We are interested in modeling  $p(y_{1:T}|x_{1:T})$. 

In the class of switching SSMs, the simplest form is the switching linear dynamical system (SLDS), where the dynamics of each regime is explained by a linear state space model. 

The discrete latent variables, denoted as $d_t \in \{1,2,\cdots,K\}$ at each time step $t= 1,2,\cdots,T$, follows a Markov chain. In particular, $d_t|d_{t-1}$  is assumed to follow a transition matrix $\Gamma \in R^{K \times K}$, where $\Gamma_{i,j} = p(d_t = j|d_{t-1}=i)$. 

The discrete latent variables $d_t$ have impact on both the continuous latent variables $z_t \in \mathbb{R}^Z$ and $y_t$ through the following  transition and emission functions:
\begin{align}
    z_{t} &= W_z^{(d_t)}z_{t-1}+W_x^{(d_t)}x_{t}+b_z^{(d_t)} + e_t,  e_t \sim N(0,\mathbf{\Sigma}_z^{(d_t)}) \label{transition}\\
    y_{t} &=  W_y^{(d_t)}z_{t}+b_y^{(d_t)} +\epsilon_t, \epsilon_t \sim N(0,\mathbf{\Sigma}_y^{(d_t)}) \label{emission}
\end{align}
where  $W_z^{(d_t)} \in \mathbb{R}^{Z\times Z}, W_x^{(d_t)} \in \mathbb{R}^{Z\times U},b_z^{(d_t)} \in \mathbb{R}^{Z\times 1}$, $\mathbf{\Sigma}_z^{(d_t)} \in  \mathbb{R}^{Z\times Z}$, $W_y^{(d_t)} \in \mathbb{R}^{D\times Z}, b_y^{(d_t)} \in \mathbb{R}^{D\times 1}$ and $\mathbf{\Sigma}_y^{(d_t)} \in \mathbb{R}^{D\times D}$.  The \textbf{transition} function in (\ref{transition}) determines the evolution of the latent variable. The \textbf{emission} function in (\ref{emission}) specifies the dynamics of the observed time series, given the state of the latent variables at time $t$.  The transition noise $e_t$ and emission noise $\epsilon_t$ are Gaussian distributed. When $K  = 1$, the model is also termed as the  Linear Gaussian State Space Model (LGSSM).

There have been several extensions to the SLDS by introducing nonlinear structures. The RSSSM is proposed in \cite{chow2013nonlinear} which adopts a pre-specified nonlinear transition function. The SVAE model \citep{johnson2016composing}  parametrizes the emission function by neural networks, while the transition function remains linear. The SNLDS model \citep{dong2020collapsed} parametrizes both the emission and transition functions with nonlinear neural networks.  The DSARF \citep{farnoosh2020deep} approximates high-dimensional time series with a multiplication of latent factors and latent weights, where the latent weights are modeled by a nonlinear autoregressive model, switched by a Markov chain of discrete latent variables. Most of the above-mentioned work assumed that the $d_t$ only influences the transition of $z_t$. Our work, on the other hand, assumes that the $d_t$ which represents the long-term dynamic changes of the time series,  affects both $y_t$ and $z_t$.

The discrete switching variables in the SLDS are assumed to be Markov, i.e. $d_t$ depends on $d_{t-1}$ only. The recurrent SLDS (rSLDS)  proposed in \cite{linderman2017bayesian} and \cite{becker2019switching}  extends the open-loop Markov dynamics and makes $d_t$ depending on the hidden state $z_{t-1}$. In \cite{nassar2019tree}, a tree structure prior is imposed on the switching variables of rSLDS, where the dynamics of the switching variables behave similarly in the same subtrees. 
The deep Rao-Blackwellised Particle Filter proposed in \cite{kurle2020deep} also allows $d_t$ to depend on $z_{t-1}$.
The SNLDS model \citep{dong2020collapsed} extends the open-loop Markov dynamics by making $d_t$ depend on the last observations. 
Such recurrent structures serve as a presence of disturbance to the switching dynamics.
Although such recurrent structures sometimes can improve the accuracy, they can also lead to unnecessarily frequent state shifts in the estimated discrete latent variables, making interpretations difficult, see our simulated toy example. 
There is a need for a well-designed architecture that allows the disturbance to be reasonably represented, and simultaneously does not lead to over frequent switching. 
In this work, we stick to a Markov dynamic of the discrete latent variable and push the non-Markov dynamics into the continuous latent variables that depend on the hidden states of a recurrent neural network summarizing the information coming from the past.  


\section{Deep Switching State Space Model (DS$^3$M)} \label{sec:formulation}
In this section, we will detail the generative and inference network of the DS$^3$M. 

\subsection{Model}
\begin{figure*}[t]
    \centering
    \begin{subfigure}[b]{0.42\textwidth}
        \centering
        \includegraphics[width=\textwidth]{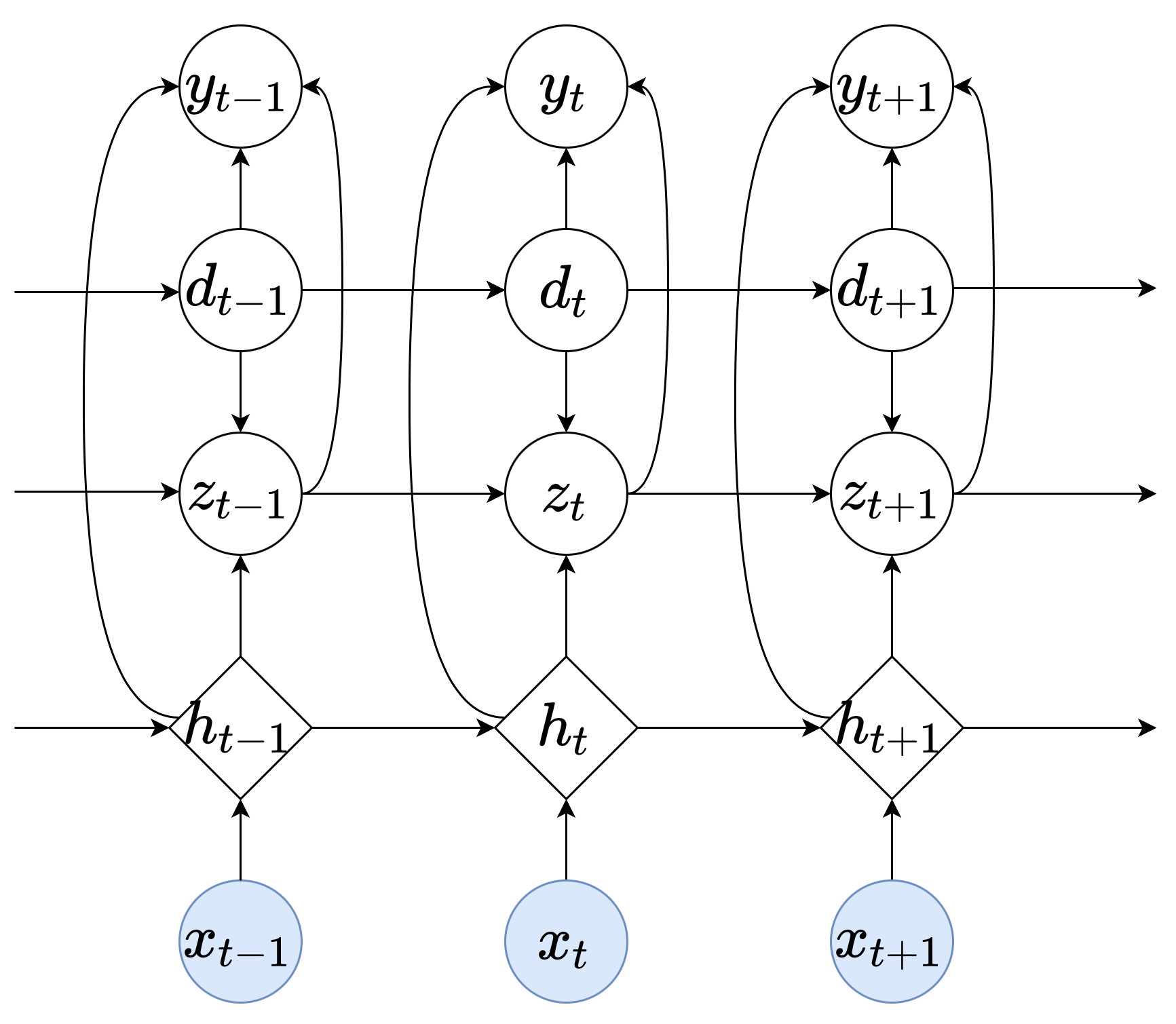}
        \caption{Generative network}
        \label{fig:generative}
    \end{subfigure}
    \hfill
    \begin{subfigure}[b]{0.45\textwidth}
        \centering
        \includegraphics[width=\textwidth]{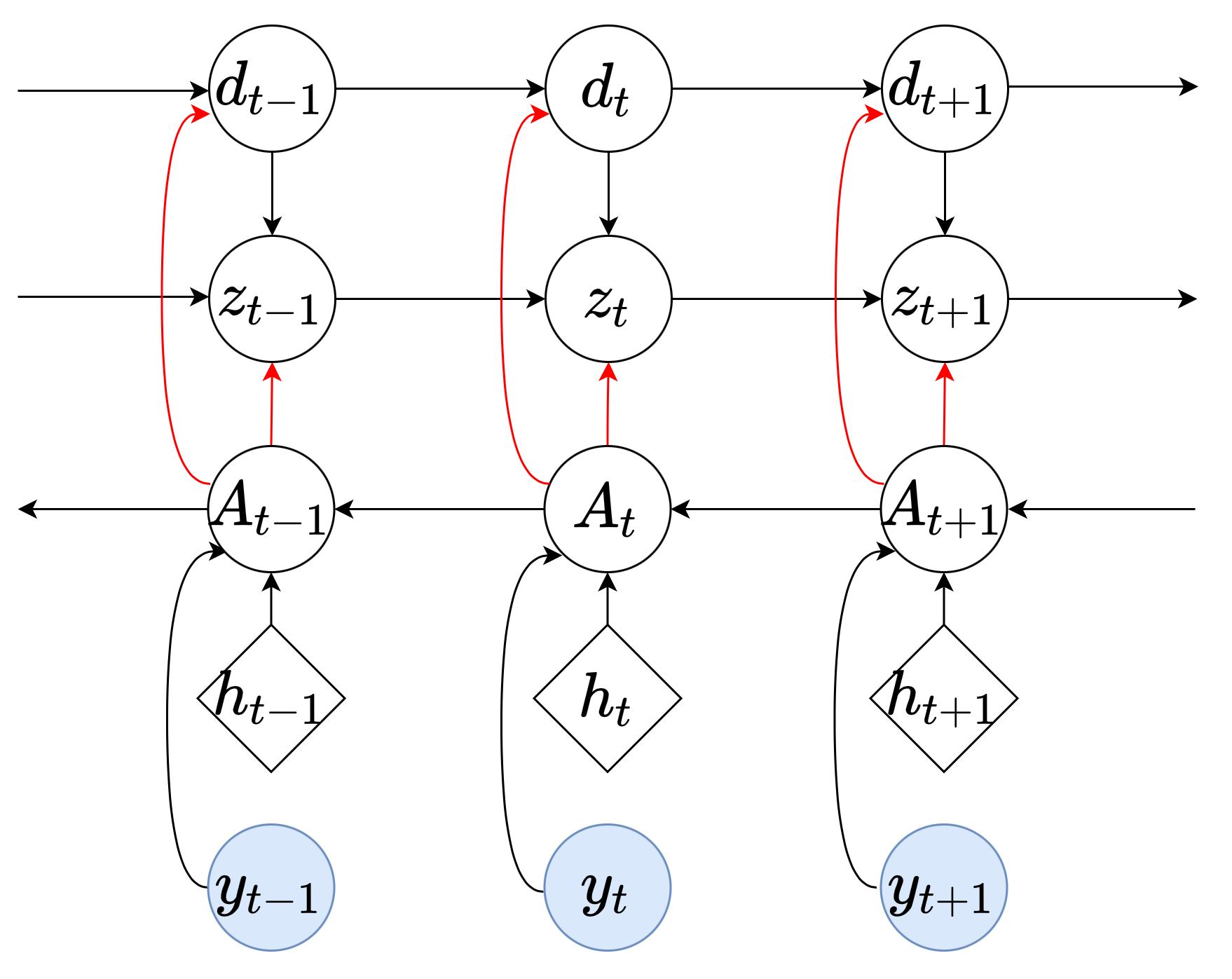}
        \caption{Inference network}
        \label{fig:inference}
    \end{subfigure}
    \caption{Deep Switching State Space Model (DS$^3$M)}
    \label{fig:dsssm}
\end{figure*}

The generating model of the DS$^3$M is as follows:
\begin{enumerate}
\item[(a)]  At time step $t$, a forward RNN is used to process the input data:
\begin{equation}
h_t = f_h(h_{t-1},x_{t})
\end{equation}
where $h_t$ is the RNN hidden state and $f_h$ can be an LSTM or GRU. 

\item[(b)]  The discrete latent variable $d_t \in \{1,\cdots,K\}$ evolves following a Markov transition matrix $\Gamma \in R^{K \times K}$ with $\Gamma_{i,j} = p(d_t = j|d_{t-1}=i)$. Here $K$ refers to the number of regime states. 

\item[(c)] {The \textbf{transition function }of the continuous latent variable $z_t \in \mathbb{R}^Z$ depends on $z _ { t - 1 } , h_t, d_t$  and is modeled as follows:
\begin{align}
    z_t  & = \mu_ {z, t}^{({d_t})}  + e_t, e_t \sim N(0, \mathbf{\mathbf{\Sigma}}_{z,t}^{({d_t})} )\\
    \mu_ {z, t}^{({d_t})}  & = f^{({d_t})} _ { z,1 } \left(  z  _ { t - 1 } ,  h  _ { t } \right) , \quad \log \mathbf { \Sigma_{z,t} }^{({d_t})} _ { t }  = f ^{({d_t})}_ {z,2} \left( z  _ { t - 1 } , h  _ { t } \right), \label{equ:ds3m_transition}
\end{align}
where the mean $\mu^{({d_t})}_{z,t} \in \mathbb{R}^Z$ and the diagonal covariance matrix $\mathbf{\Sigma}^{({d_t})}_{z,t} = \text{Diag}(\sigma_1^{d_t}, \sigma_2^{d_t}, \cdots, \sigma_Z^{d_t})$ are modeled with neural networks $f^{({d_t})}_{z,1}$ and $f^{({d_t})}_{z,2}$. In fact, $z_t\sim N(\mu_ {z, t}^{({d_t})}, \mathbf{\mathbf{\Sigma}}_{z,t}^{({d_t})} )$.}

\item[(d)] The \textbf{emission function} for the time series $y_t$  depends on $z _ { t  } , h_t,d_t$  and is modeled as follows:
\begin{align}
y_t  & = \mu_ {y, t}^{({d_t})}  + \epsilon_t, \epsilon_t \sim N(0, \mathbf{\Sigma}_{y,t}^{({d_t})} )\\
\mu_{y,t}^{({d_t})}  & = f^{({d_t})} _ { y, 1 } \left(  z  _ { t } ,  h  _ { t } \right) , \quad \log \mathbf { \Sigma }^{({d_t})} _ {y,t}  = f ^{({d_t})}_ {y,2} \left( z  _ { t } , h  _ { t } \right),  \label{equ:ds3m_emission}
\end{align}
where the mean $\mu^{({d_t})}_{y,t} \in \mathbb{R}^D$ and the  covariance matrix $\mathbf{\Sigma}^{({d_t})}_{y,t} \in \mathbb{R}^{D \times D}$ are modeled with neural networks $f^{({d_t})}_{y,1}$ and $f^{({d_t})}_{y,2}$. In fact, $y_t\sim N(\mu_ {y, t}^{({d_t})}, \mathbf{\mathbf{\Sigma}}_{y,t}^{({d_t})} )$.
\end{enumerate}
Figure \ref{fig:generative} displays a graphical representation of the DS$^3$M. 

The  DS$^3$M differs from the SLDS model reviewed in Section \ref{sec:background} in several aspects: (1) DS$^3$M  introduced an RNN model to encode all the historical information in each time step. (2) The transition and the emission function become nonlinear and  depend on the hidden states of the RNN.

It is also important to stress the key differences between DS$^3$M and the state-of-the-art SNLDS. We stack an RNN below the SSSM and design a direct connection of the hidden state $h_t$ to the time series $y_t$ inspired by the skip connection in ResNet \citep{he2016deep}, Transformers \citep{vaswani2017attention} and SRNN \citep{fraccaro2016sequential}.  While in SNLDS, there is no RNN and no hidden state $h_t$. Compared Equation \eqref{equ:snlds_transition} with  \eqref{equ:ds3m_transition},  Equation \eqref{equ:snlds_emission} with  \eqref{equ:ds3m_emission}, the transition and emission functions of the SNLDS do not depend on $h_t$ and can be represented as 
\begin{align}
z_t  & = \mu_ {z, t}^{({d_t})}  + e_t, e_t \sim N(0, \mathbf{\Sigma}_{z,t}^{({d_t})} )\\
\mu_ {z, t}^{({d_t})}  & = f^{({d_t})} _ { z,1 } \left(  z  _ { t - 1 } \right) , \quad \log \mathbf { \Sigma_{z,t} }^{({d_t})} _ { t }  = f ^{({d_t})}_ {z,2} \left( z  _ { t - 1 }  \right), \label{equ:snlds_transition}
\end{align}
\begin{align}
y_t  & = \mu_ {y, t}^{({d_t})}  + \epsilon_t, \epsilon_t \sim N(0, \mathbf{\Sigma}_{y,t}^{({d_t})} )\\
\mu_{y,t}^{({d_t})}  & = f^{({d_t})} _ { y, 1 } \left(  z  _ { t }  \right) , \quad \log \mathbf { \Sigma }^{({d_t})} _ {y,t}  = f ^{({d_t})}_ {y,2} \left( z  _ { t } \right).   \label{equ:snlds_emission}
\end{align}
From a modeling aspect, a lack of this connection will force the continuous latent variable $z_t$ to encode all the relevant historical information. The connection between $y_t$ and $h_t$ on the other hand allows a clear structure, where both the deterministic hidden states and the stochastic latent variables can separately encode different aspects of information. The addition of the RNN and the skip connection is important as we aim at the out-of-sample prediction task, while the focus of SNLDS is the segmentation of time series, i.e. identify the regimes (in-sample inference).

    \begin{remark}
        In the emission function, we assume that the emission distribution for the time series $y_t$ is a Gaussian distribution for simplicity. The emission distribution can be extended to other parametric distributions. For example, one could choose the emission distribution to be lognormal for asymmetric and positive time series and model the mean and variance of the lognormal distribution using neural networks. 
\end{remark}

\subsection{Estimation of parameters}
Denote $\theta = \{f_h,\Gamma,\{f_{z,1}^{(k)}\}_{k=1}^{K},\{f_{z,2}^{(k)}\}_{k=1}^{K},\{f_{y,1}^{(k)}\}_{k=1}^{K},\{f_{y,2}^{(k)}\}_{k=1}^{K}\}$ as the parameters of the DS$^3$M. The joint probability is represented as
\begin{equation}
    \begin{aligned} 
        &p_{\theta}\left(\mathbf{y}_{1 : T}, \mathbf{z}_{1 : T}, \mathbf{d}_{1 : T} | \mathbf{x}_{1 : T}\right) \\
        &=\textstyle \prod_{t=1}^{T} p_{\theta}\left(\mathbf{y}_{t} | \mathbf{z}_{t}, \mathbf{h}_{t},\mathbf{d}_{t}\right) p_{\theta}\left(\mathbf{z}_{t} | \mathbf{z}_{t-1}, \mathbf{h}_{t}, \mathbf{d}_{t}\right) p_{\theta}\left(\mathbf{d}_{t} | \mathbf{d}_{t-1}\right).
    \end{aligned}
    \label{equ:jointprobability}
\end{equation}
To obtain the likeliood of observations, denoted as  $\mathcal{L}_{}(\theta) = p_{\theta}\left(\mathbf{y}_{1 : T}| \mathbf{x}_{1 : T}\right)$, it requires to marginal out  $\mathbf{z}_{1 : T}$ and $\mathbf{d}_{1 : T}$ in (\ref{equ:jointprobability}). However, due to the non-linearity introduced by neural networks, it is intractable to do the marginalization and the maximum likelihood method is not applicable for the estimation of parameters in this setting.

In order to conduct the estimation of parameters, we develop a scalable learning and inference algorithm for the DS$^3$M using variational inference instead. Specifically, we approximate the  posterior  $p(\mathbf{z}_{1 : T},\mathbf{d}_{1 : T}|\mathbf{x}_{1 : T},\mathbf{y}_{1 : T})$  with an appximated posterior $q_{\phi}\left(\mathbf{z}_{1 : T},\mathbf{d}_{1 : T}| \mathbf{y}_{1 : T},\mathbf{x}_{1 : T}\right)$ with parameter $\phi$, and then maximize an evidence lower bound $\mbox{ELBO}(\theta,\phi) \leq 	\mathcal{L}(\theta)$ with respect to both $\theta$ and $\phi$ to obtain the parameters $\theta$.  

\subsubsection{The ELBO and the approximated posterior}
The $\mbox{ELBO}(\theta,\phi)$ can be calculated as follows:
\begin{equation}\label{equ:ELBO}
        \begin{aligned}
         \mbox{ELBO}(\theta,\phi) &=
            \mathbb{E}_{q_{\phi}}\left[\log p_{\theta}\left(\mathbf{y}_{1 : T} | \mathbf{z}_{1 : T},\mathbf{d}_{1 : T}, \mathbf{x}_{1 : T}\right)\right] \\
            &\quad -\mathrm{KL}\left(q_{\phi}\left(\mathbf{z}_{1 : T},\mathbf{d}_{1 : T}| \mathbf{y}_{1 : T},\mathbf{x}_{1 : T}\right) \| p_{\theta}\left(\mathbf{z}_{1 : T}, \mathbf{d}_{1 : T}| \mathbf{x}_{1 : T} \right)\right).
        \end{aligned}
\end{equation}
The \mbox{ELBO} is tight, i.e. $\mathcal{L}(\theta) = \mbox{ELBO}(\theta,\phi)$, only when the approximated posterior $q_{\phi}\left(\mathbf{z}_{1 : T},\mathbf{d}_{1 : T}| \mathbf{y}_{1 : T},\mathbf{x}_{1 : T}\right) $ is equal to the true posterior $p_{\theta}\left(\mathbf{z}_{1 : T}, \mathbf{d}_{1 : T}| \mathbf{y}_{1 : T},\mathbf{x}_{1 : T} \right)$, which is unfortunately intractable.  Nevertheless, we aim to design an approximated posterior $q_{\phi}\left(\mathbf{z}_{1 : T},\mathbf{d}_{1 : T}| \mathbf{y}_{1 : T},\mathbf{x}_{1 : T}\right)$ that could mimic the dynamics of the true posterior, so that we could achieve a tighter $\mbox{ELBO}$. According to  the d-separation \citep{geiger1990identifying} of the generative network, the true posterior can be factorized in the following form:
\begin{equation}
    \begin{aligned}
        &\quad p_{\theta}\left(\mathbf{z}_{1 : T} , \mathbf{d}_{1 : T}| \mathbf{y}_{1 : T}, \mathbf{x}_{1 : T}\right)\\
        &= \textstyle \prod_{t} p_{\theta}\left(\mathbf{z}_{t} | \mathbf{z}_{t-1}, \mathbf{d}_{t}, \mathbf{y}_{t : T},\mathbf{h}_{t : T}\right)p_{\theta}\left(\mathbf{d}_{t} | \mathbf{d}_{t-1}, \mathbf{y}_{t : T},\mathbf{h}_{t : T}\right),
    \end{aligned}
\end{equation}
where the posterior of $\mathbf{z_t},\mathbf{d_t}$ depends on the past information encoded in $\{\mathbf{z}_{t-1},\mathbf{d}_{t-1}\}$ as well as the future information in $\{\mathbf{y}_{t : T},\mathbf{h}_{t : T}\}$. Therefore, we design the approximated posterior to have the same dependency as the true posterior as follows:
\begin{align}
    &\quad q_{\phi}\left(\mathbf{z}_{1 : T} , \mathbf{d}_{1 : T}| \mathbf{y}_{1 : T}, \mathbf{x}_{1 : T}\right) \notag \\
    &=\textstyle\prod_{t} q_{\phi_z}\left(\mathbf{z}_{t} | \mathbf{z}_{t-1}, \mathbf{d}_{t}, \mathbf{A}_t\right)q	_{\phi_d}\left(\mathbf{d}_{t} | \mathbf{d}_{t-1}, \mathbf{A}_t\right),
    \label{equ:inference}
\end{align}
where $\mathbf{A}_{t}=g_{\phi_{\mathrm{A}}}\left(\mathbf{A}_{t+1},\left[\mathbf{y}_{t}, \mathbf{h}_{t}\right]\right)$ and $\phi = \{\phi_z,\phi_d,\phi_A\}$.  We parameterize  $g_{\phi_{\mathrm{A}}}$ as a backward RNN and $q_{\phi_z}\left(\mathbf{z}_{t} | \mathbf{z}_{t-1}, \mathbf{d}_{t}, \mathbf{A}_t\right)$ with a Gaussian probabilistic density:
\begin{align}
    &\mathbf{z}_{t} = 	\mu ^{({d_t})}_ {z, t } + e_t,  e_t \sim N(0,1) \label{equ:q_z_transition_1}\\
    &	\mu ^{({d_t})}_ {z, t }  = g^{({d_t})} _ { 1 } \left(  z  _ { t - 1 } ,  \mathbf{A}_t\right) , \quad \log \mathbf{\Sigma}^{({d_t})} _ {z, t }  = g ^{({d_t})}_ {2} \left( z  _ { t - 1 } , \mathbf{A}_t\right) \label{equ:q_z_transition_2},
\end{align}
where $\mathbf{z} _ { t }$ is a Gaussian variable with mean $\mu^{({d_t})}_{z,t}$ and diagonal variance matrix $\mathbf{\Sigma}^{({d_t})}_{z,t}$  determined by neural network models $g^{({d_t})}_{1}$ and $g^{({d_t})}_{2}$. $z_1 \sim N(0,1) $.  Moreover, $q_{\phi_d}\left(\mathbf{d}_{t} | \mathbf{A}_t,\mathbf{d}_{t-1}\right)$ is parameterized as a Categorical distribution:
\begin{equation}
d_t \sim \text{Cat}(\text{softmax}(W^{({d_{t-1}})}\mathbf{A}_t)),
\label{equ: q_discrete} 
\end{equation}
and $d_1 \sim \text{Uniform}(1,2, \cdots, K) $. The graphical model of the inference network is shown in Figure \ref{fig:inference}.
\subsubsection{Factorization and approximation of the ELBO}
According to the conditional independence inherited in the DS$^3$M, the conditional probability of the latent variables on the observations can be factorized in the following form:
\begin{equation}\label{equ:prior}
        p_{\theta}\left(\mathbf{z}_{1 : T}, \mathbf{d}_{1 : T}| \mathbf{x}_{1 : T} \right) 
        =\textstyle\prod_{t}p_{\theta}(\mathbf{d}_{t}|\mathbf{d}_{t-1})p_{\theta}\left(\mathbf{z}_{t}|\mathbf{z}_{t-1},\mathbf{d}_{t}, \mathbf{h}_{t} \right).
\end{equation}

With the  approximate posterior in Equation \eqref{equ:inference} and the factorization in \eqref{equ:prior}, the \mbox{ELBO} can be derived as follows:
\begin{equation}
        \begin{aligned} 
        \mbox{ELBO}(\theta,\phi)
            &=\textstyle \sum_{t}\biggl\{\mathbb{E}_{q_{\phi}^*(\mathbf{z}_{t-1},\mathbf{d}_{t-1})}\sum_{\mathbf{d}_t}{q_{\phi_d}(\mathbf{d}_{t})}\mathbb{E}_{q_{\phi_z}(\mathbf{z}_{t})}\left[\log  p_{\theta}\left(\mathbf{y}_{t} | \mathbf{z}_{t},  \mathbf{d}_t , \mathbf{h}_{t}\right)\right]  -\\
            &\textstyle \mathbb{E}_{q_{\phi}^{*}\left(\mathbf{z}_{t-1},d_{t-1}\right)}\sum_{\mathbf{d}_t}{q_{\phi_d}\left(\mathbf{d}_{t} \right)} KL\left[ q_{\phi_z}\left(\mathbf{z}_{t} | \mathbf{z}_{t-1}, \mathbf{d}_{t}, \mathbf{A}_t\right)\| p_{\theta}\left(\mathbf{z}_{t}|\mathbf{z}_{t-1},\mathbf{d}_{t}, \mathbf{h}_{t} \right)\right] \\
            & \textstyle    -\mathbb{E}_{q_{\phi}^{*}\left(d_{t-2}\right)}\sum_{\mathbf{d}_{t-1}}{q_{\phi_d}\left(\mathbf{d}_{t-1} \right)}  KL\left[ q_{\phi_d}\left(\mathbf{d}_{t} | \mathbf{d}_{t-1}, \mathbf{A}_t\right)\| p_{\theta}(\mathbf{d}_{t}|\mathbf{d}_{t-1})\right]\biggl\},
        \end{aligned}
    \label{equ:elbo_factorized}
\end{equation}
where
$q_{\phi}^{*}\left(\mathbf{z}_{t},\mathbf{d}_{t}\right)=\int q_{\phi}\left(\mathbf{z}_{1 : t},\mathbf{d}_{1 : t} |\mathbf{y}_{1 : T}, \mathbf{x}_{1:T}\right) \mathrm{d} \mathbf{z}_{1 : t-1}\mathrm{d} \mathbf{d}_{1 : t-1}$ and\\ $q_{\phi}^{*}\left(\mathbf{d}_{t}\right) =\int q_{\phi_d}\left(\mathbf{d}_{1:t} | \mathbf{y}_{1 : T}, \mathbf{x}_{1:T}\right)\mathrm{d}  \mathbf{d}_{1: t-1}$.  The detailed derivation can be found in Appendix \ref{appen: elbo}.

We can approximate the \mbox{ELBO} in \eqref{equ:elbo_factorized} using a Monte Carlo method. We need to obtain samples $(z_t^{(s)},d_t^{(s)}) \text{ for } t= 1 \cdots, T$ from $q_{\phi}^{*}\left(\mathbf{z}_{t},\mathbf{d}_{t}\right)$ using ancestral sampling according to Equation \eqref{equ:inference}. Given a sample $(\mathbf{z}_{t-1}^{(s)},\mathbf{d}_{t-1}^{(s)})$ from $q_{\phi}^{*}\left(\mathbf{z}_{t-1},\mathbf{d}_{t-1}\right)$, a sample $(\mathbf{z}_{t}^{(s)},\mathbf{d}_{t}^{(s)})$ from $q_{\phi}\left(\mathbf{z}_{t},\mathbf{d}_{t} |\mathbf{z}_{t-1}^{(s)},\mathbf{d}_{t-1}^{(s)},\mathbf{y}_{t : T}, \mathbf{h}_{t:T}\right)$ will follow $q_{\phi}^{*}\left(\mathbf{z}_{t},\mathbf{d}_{t}\right)$.  Therefore, we could adopt the following procedures to obtain the Monte Carlo samples $(z_t^{(s)},d_t^{(s)}) \text{ for } t= 1 \cdots, T$ :
\begin{enumerate}
    \item[i.] Sample $d_1^{(s)} \sim  \text{Uniform}(1,2, \cdots, K)$, 
    sample $z_1^{(s)} \sim N(0,1)$ 
    \item[ii.]For $t = 2, \cdots, T$ , sample $d_t^{(s)}$ according to the density $q_{\phi_d}\left(\mathbf{d}_{t} | \mathbf{A}_t,\mathbf{d}_{t-1}\right)$ based on  Equation \eqref{equ: q_discrete}; sample $z_t^{(s)}$ according to  the density $q_{\phi_z}\left(\mathbf{z}_1 | \mathbf{z}_{t-1}, \mathbf{d}_{t}, \mathbf{A}_t\right)$ based on Equation \eqref{equ:q_z_transition_1} and \eqref{equ:q_z_transition_2}.
\end{enumerate}

The approximated  \mbox{ELBO} can then be represented as follows:
\begin{equation}
\begin{aligned}
\mbox{ELBO}(\theta,\phi)
    &\textstyle \approx\sum_{t}\biggl\{\sum_{\mathbf{d}_{t}}{q_{\phi_d}\left(\mathbf{d}_{t} \right)}\log  p_{\theta}\left(\mathbf{y}_{t} | \mathbf{z}_{t}^{(s)}, \mathbf{d}_{t}, \mathbf{h}_{t}\right) \\
    &\textstyle \quad - \sum_{\mathbf{d}_{t}}{q_{\phi_d}\left(\mathbf{d}_{t} \right)} KL\left[ q_{\phi_z}\left(\mathbf{z}_{t} | \mathbf{z}_{t-1}^{(s)}, \mathbf{d}_{t}, \mathbf{A}_t\right)\| p_{\theta}\left(\mathbf{z}_{t}|\mathbf{z}_{t-1}^{(s)},\mathbf{d}_{t}, \mathbf{h}_{t} \right)\right] \\
    &  \textstyle \quad -\sum_{\mathbf{d}_{t-1}} {q_{\phi_d}\left(\mathbf{d}_{t-1} \right)} KL\left[ q_{\phi_d}\left(\mathbf{d}_{t} | \mathbf{d}_{t-1}, \mathbf{A}_t\right)\| p_{\theta}(\mathbf{d}_{t}|\mathbf{d}_{t-1})\right]\biggr\}.
\end{aligned}	
\label{equ:approximatedELBO}
\end{equation}

\subsubsection{Gradients of the ELBO}
We use the gradient descent algorithm to optimize the approximated ELBO in \eqref{equ:approximatedELBO}. While it is easy to obtain the gradient of ELBO with respect to $\theta$ ($\nabla_{\theta} \mbox{ELBO}(\theta,\phi)$), it is not the case for the gradient of ELBO with respect to $\phi$  ($ \nabla_{\phi}\mbox{ELBO}(\theta,\phi))$. 
The score function gradient estimator \citep{williams1992simple} suffers from high variance and the reparameterization approach is often used to reduce the variance \citep{rezende2014stochastic,kingma2013auto}.

For the continuous latent variable $z_t$ in the \mbox{ELBO}, we apply the reparameterization approach. 
Specifically, we generate a sample $e_t \sim N(0,1)$, and then use  $\mu_t + \epsilon_t \mathbf{\Sigma}_t$ as a sample of $z_t \sim N(\mu_t,\mathbf{\Sigma}_t)$ in the above Monte Carlo approximation.  By adopting this approach, 
the gradients can then be backpropagated through the continuous random variables.

The reparameterization trick, however, is not applicable for the discrete random variable $d_t$.  One commonly used reparameterization trick for discrete random variables is the Gumble-softmax reparameterization trick \citep{dong2020collapsed}. However, this will result in non-interger sampling values for  $d_t$, which are invalid for our model.   As an alternative approach,  in \eqref{equ:approximatedELBO}, we marginalize out the discrete variable $d_t$ with a summation over its probability at each time step $t$, and do not marginalize out the discrete variable before time $t$. This introduces a biased gradient estimator for $\phi_d$ and $\phi_z$ and it can be viewed as gradient clips where the gradients from the previous time steps are ignored. In our experiments, such an approximation performs very well compared to the unbiased score function estimator and we consider the bias negligible. We have done some experiments that marginalized more than one step in  \eqref{equ:approximatedELBO}. The performance is similar, but the time complexity in computing \eqref{equ:approximatedELBO}  will increase from $O(KT)$ to $O(K^2T)$. Therefore, we stick to the current setting.

It is worth mentioning that the SNLDS marginalizes the discrete latent variables using the exact posterior derived with the forward-backward algorithm, while the DS$^3$M marginalizes the discrete latent variables using the approximate posterior at each time step. One potential problem of marginalizing the discrete latent variables using the exact posterior is that the approximate posterior for $z_t$ does not depend on $d_t$ anymore. This could lead to a severe posterior collapse problem that $d_t$ is not used at all. They proposed an entropy regularizer to encourage an evenly distributed posterior for $d_t$. However, there is no guarantee that an evenly distributed posterior will produce a meaningful interpretation. In contrast, we use an approximated posterior of $z_t$ that depends on $d_t$ to form a connected inference between $z_t$  and $d_t$. The posterior collapse problem for the discrete latent variables does not appear in our experiments. 

A summary of the structured inference algorithm is given in Algorithm \ref{alg:DS$^3$M}.
\begin{algorithm}[H]\small
    \caption{Structured Inference Algorithm for DS$^3$M}
    \label{alg:DS$^3$M}
    \begin{algorithmic}
        \STATE \textbf{Inputs}: $\{x_{1:T}\}_{i=1}^N$, $\{y_{1:T}\}_{i=1}^N$, randomly initialized $\phi^{(0)}$ and $\theta^{(0)}$\\
        \STATE \textbf{Outputs}: $\theta$, $\phi$
        \WHILE{$Iter$ $<$ M} 
        \STATE1. Sample a mini-batch sequences $\{x_{1:T}\}_{i=1}^B$, $\{y_{1:T}\}_{i=1}^B$  from the dataset
        \STATE2. Generate  $z_t^{(s)}, d_t^{(s)}$ for $t=1,2,\cdots,T$ sequentially according to  (\ref{equ:inference}) to approximate the \mbox{ELBO} in  (\ref{equ:approximatedELBO})
        \STATE3. Derive  $\nabla_{\theta} \mbox{ELBO}(\theta,\phi)$ and $\nabla_{\phi}  \mbox{ELBO}(\theta,\phi)$
        \STATE4. Update $\theta^{(Iter)}, \phi^{(Iter)}$ using the ADAM, set $Iter = Iter+1$
        \ENDWHILE
    \end{algorithmic}
\end{algorithm}

\begin{remark}
    In our paper, we assume the number of switching statuses to be known. It is possible to use a Bayesian mixture modeling approach \citep{giudici2003mixtures} to automatically determine the number of switch statuses in our $DS^3M$ model. However, adapting our model to this non-parametric Bayesian framework is not trivial. It would require joint modeling of the number of switch statuses along with the parameters of the $DS^3M$, as well as a redesign of the estimation algorithm of the parameters. Therefore, we leave this for future research. 
    
    {\color{blue} 
    Both underestimation and overestimation of the number of switching statuses can lead to suboptimal forecasting performance. Specifically, underestimating the regimes may result in modeling bias and reduced adaptability to regime shifts, while overestimating can increase the variance in predictions due to overfitting, thereby reducing the robustness of forecasts.

    To mitigate this issue, we suggest employing data-driven methods, such as cross-validation, to select the optimal number of states. Additionally, extending likelihood-based tests \citep{giudici2000likelihood} to this context could be another viable approach. We consider this a direction for future research.
    }
\end{remark}

\subsection{Predictive distributions}
Given a trained model, we are interested in the predictive distributions for the one-step-ahead to $\tau$-step-ahead observations $\{y_{T+1}, \cdots, y_{T+\tau}\}$ and the discrete latent variables $\{d_{T+1}, \cdots, d_{T+\tau}\}$. We first make inference on the posterior distributions of $\{z_t, d_t\}_{t=1}^T$ and then generate samples of $\{z_t^{(s)}, d_t^{(s)},y_t^{(s)}\}_{t=T}^{T +\tau}$,  $s = 1, \cdots, S$ and  $S$ represents the number of Monte Carlo samples. The predictive distributions are then approximated with the empirical distribution functions of the generated samples.
The predictive distributions can be obtained as follows:
\begin{equation}
\begin{aligned}
p(y_{T+1}|x_{1:T},y_{1:T})&\approx \textstyle \frac{1}{S}\sum_{s=1}^{S}p(y_{T+1}|z_{T+1}^{(s)},h_{T+1}), \\
p(d_{T+1}|x_{1:T},y_{1:T}) &\approx \textstyle\frac{1}{S}\sum_{s=1}^{S} p(d_{T+1}|d_{T}^{(s)}),\\
p(z_{T+1}|x_{1:T},y_{1:T})&\approx\textstyle \frac{1}{S}\sum_{s=1}^{S}p(z_{T+1}|d_{T+1}^{(s)},z_{T}^{(s)}).
\end{aligned}
\end{equation}

\subsection{Theoretical analysis of the DS$^3$M}
In the following, we demonstrate the presence of stability of DS$^3$M.

The latent continuous state in DS\textsuperscript{3}M 
$z_t \sim \mathcal{N}(\mu_{z,t}^{(d_t)}, \Sigma_{z,t}^{(d_t)})$
where the mean $\mu_{z,t}^{(d_t)}$ and the diagonal covariance matrix $\Sigma_{z,t}^{(d_t)}$ are parameterized by neural networks $f^{(d_t)}_{z,1}$ and $f^{(d_t)}_{z,2}$, respectively. Similarly, the observed time series is modeled as $y_t \sim \mathcal{N}(\mu_{y,t}^{(d_t)}, \Sigma_{y,t}^{(d_t)})$,
where where the mean $\mu_{y,t}^{(d_t)}$ and the  covariance matrix $\Sigma_{y,t}^{(d_t)}$ are parameterized by neural networks $f^{(d_t)}_{y,1}$ and $f^{(d_t)}_{y,2}$, respectively. 

\begin{theorem}\label{therom:stability}
The DS$^3$M is stable, i.e. the latent variable $z_t$ and the observed variable $y_t$ are globally mean-square stable if the 2-norms of all weight matrices and activation scaling matrices are upper bounded by 1. 
\end{theorem}

Since DS\textsuperscript{3}M employs ReLU (Rectified Linear Unit) activation functions, this naturally satisfies the constraints on activation scaling matrices. Global mean-square stability guarantees that the system states remain bounded in the mean-square sense over time, effectively preventing divergence and ensuring the reliability of long-term predictions. This property enhances the robustness and generalization capabilities of DS\textsuperscript{3}M, making it a reliable framework for capturing intricate temporal dependencies and ensuring consistent performance across diverse datasets. The proof for the Theorem \ref{therom:stability} is provided in the Appendix. 

\section{Experiments} \label{sec:simulation}
In this section, we evaluate DS$^3$M through various experiments. We first consider a simulated 1-dimensional (1-d) time series whose true dynamics follow a nonlinear switching state space model, as well as a simulated 10-dimensional (10-d) time series based on the Lorenz attractor. Furthermore, we apply DS$^3$M to several real-world datasets spanning diverse applications, such as health care, transportation, energy, and econometrics. Both simulations and real data analyses demonstrate that DS$^3$M effectively captures switching regimes and achieves competitive predictive accuracy when compared with several state-of-the-art methods, including SRNN, GRU, DSARF, and SNLDS. Specifically, SRNN can be considered our model without discrete latent variables. DSARF and SNLDS are two nonlinear dynamic latent variable models for time series that incorporate both continuous and discrete latent variables. DSARF's superior performance over models like rSLDS and SLDS in time series forecasting with similar datasets has been demonstrated \citep{farnoosh2020deep}. Hence, we omit the comparison of SLDS and rSLDS in the subsequent analysis. To ensure a fair comparison, we select the same datasets and employ the official codes of DSARF and SNLDS. Further details about hyperparameters are provided in the supplementary material.
For a fair comparison, we select the same data sets and use the official codes of DSARF and SNLDS.  Details of the hyperparameters are provided in the supplementary material.
\subsection{Simulations}
\paragraph{Toy example} 	
For the toy simulated example, we simulated data with a length of 2000 from the following nonlinear switching state space model:
\begin{equation}
    \begin{aligned}
        & d_{0} \sim \text{Bernouli}(0.5),  z_{0}  =0\\
        & d_{t} | d_{t-1} \sim \Gamma= \begin{bmatrix} 0.95&0.05\\0.05&0.95 \end{bmatrix}, d_t \in \{0,1\}\\
        &Z_{t|d_t = 0}=0.6Z_{t-1}+ 0.4\times \tanh(X_{t} + Z_{t-1})+w_{t}^{(0)}, \\
        &Z_{t|d_t = 1}=0.1 Z_{t-1}+0.2\times \sin( X_{\mathrm{t}} + Z_{t-1})+w_{t}^{(1)}, \\
        &Y_{t|d_t = 0}= 1.5 Z_{t}+ \tanh(Z_{t})+v_{t}^{(0)} ,\\
        &Y_{t|d_t = 1}= 0.5 Z_{t}+ \sin(Z_{t})+v_{t}^{(1)} , \\
        &w_{t}^{(0)} \sim N(0,10), w_{t}^{(1)} \sim N(0,1), \\
        & v_{t}^{(0)} \sim N(0,5),v_{t}^{(1)} \sim N(0,0.5)\\
    \end{aligned}
\end{equation}
For the simulated time series, the switching indicator $d_t$ controls both the dynamics of the continuous latent variable $z_t$ and the observation $y_t$. By design, $y_t$ is much more volatile (has higher variance) when $d_t = 0$ compared with $d_t = 1$. Note that we crafted the Markovian transition matrix with the intention of maintaining a high probability for the regime to remain in its current state, rather than undergoing frequent and chaotic shifts. This deliberate design choice reflects the characteristics often observed in various real-world contexts. Our aim is to closely mimic the realistic settings of these fields, where the relative stability of regimes is a prevalent feature.
We transform the time series into subsequences with a length of 20, resulting in 1980 subsequences.  The first 1000, the following 480, and the last 500 subsequences are used for training, validation, and testing.  We set $x_t = y_{t-1}$.	

\begin{figure*}
    \centering
    \begin{subfigure}{0.58\textwidth}
        \includegraphics[width=1\linewidth]{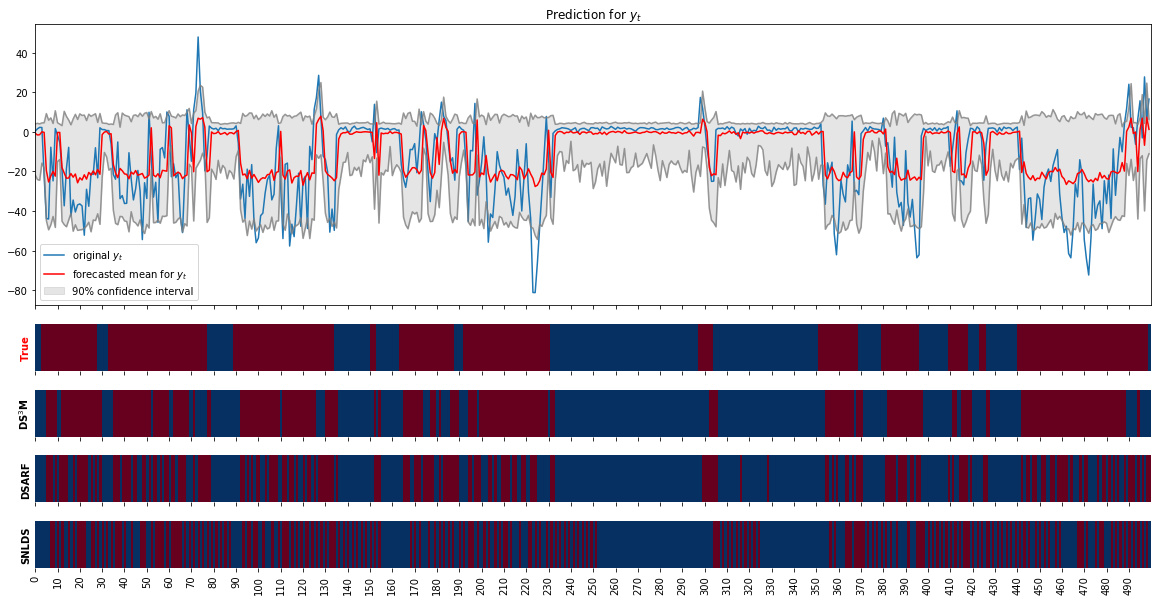}
        \caption{Prediction for the toy example. The red  color means $d_t =0$ and the blue color means $d_t  = 1$}
        \label{fig:toypredict}
    \end{subfigure}
    \begin{subfigure}{0.41\textwidth}
        \centering
        \includegraphics[width=1.1\linewidth]{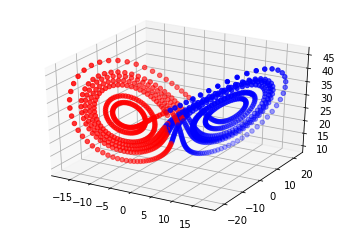}
        \caption{The forecasted switching variable against the true $z_t$.  }
        \label{fig:lorenzswitching}
    \end{subfigure}
    \caption{Plots for the two simulated datasets}
\end{figure*}

Figure \ref{fig:toypredict} showcases the one-step-ahead forecasting results (one experiment run) of DS$^3$M for the testing data, along with the predicted switching indicators of DS$^3$M, SNLDS, and DSARF. Notably, the predictive means of the observations closely align with the actual observations, while the 90\% confidence intervals effectively encompass a majority of the data points. Furthermore, DS$^3$M adeptly adapts by offering wider confidence intervals during volatile periods and narrower ones during more stable data phases. Importantly, the learned transition matrix, $\begin{bmatrix} 0.91 & 0.09 \\ 0.18 & 0.82 \end{bmatrix}$, exhibits close alignment with the true transition matrix.

Table \ref{tab:simulation} presents a summary of forecasting and inference accuracy across five experiment runs. DS$^3$M excels with lower forecasting RMSE (Root Mean Square Error) for observations, showing a relative enhancement of 11.46\% and 4.41\% over SNLDS and DSARF respectively. Furthermore, DS$^3$M achieves significantly higher state prediction accuracy (with a relative improvement of 44.99\% compared to SNLDS) and a higher F1 score (with a relative improvement of 41.77\% compared to SNLDS) for the switching indicators. While DSARF and DS$^3$M exhibit comparable state prediction accuracy and F1 score, Figure \ref{fig:toypredict} shows that DS$^3$M provides much more reliable predictions for switching indicators as compared to the ground truth, while SNLDS and DSARF tend to switch too frequently. The mean duration lengths of the two states in DS$^3$M are 7.509 and 7.634, although they are still smaller than the true values (24 and 24). However, this performance is notably better than the alternatives, where duration lengths range around 1--4. When applied to segmenting time series (inference), DS$^3$M also showcases superior accuracy and F1 scores compared to SNLDS, while performing similarly to DSARF.

\begin{table*}[t]
    \centering
    \caption{Summary of the simulation results (mean $\pm$ standard deviation) over five experiment runs}
    \resizebox{\textwidth}{!}{  
        \begin{tabular}{rlrrrrrr}
            \toprule
            &       & \multicolumn{3}{c}{Toy} & \multicolumn{3}{c}{Lorenz} \\
            \cmidrule{3-8}          &       & DS$^3$M  & SNLDS & DSARF & DS$^3$M  & SNLDS & DSARF \\
            \midrule
            \multicolumn{1}{l}{Forecasting} & RMSE  & \textbf{14.572} $\pm$ 0.352 & 16.541 $\pm$ 0.024 & 15.244 $\pm$ 0.136 & 0.168 $\pm$ 0.017 & 0.226 $\pm$ 0.065 & \textbf{0.030} $\pm$ 0.000 \\
            & Duration for dt=1 & \textbf{7.509} $\pm$ 1.579 & 1.282 $\pm$ 0.001 & 3.946 $\pm$ 0.426 & -     & -     & - \\
            & Duration for dt=0 & \textbf{7.634} $\pm$ 1.667 & 1.667 $\pm$ 0.012 & 3.274 $\pm$ 0.985 & -     & -     & - \\
            & Accuracy (\%) & \textbf{0.788} $\pm$ 0.033 & 0.543 $\pm$ 0.001 & 0.765 $\pm$ 0.047 & \textbf{0.882} $\pm$ 0.079 & 0.616 $\pm$ 0.065 & 0.788 $\pm$ 0.143 \\
            & F1 score & \textbf{0.778} $\pm$ 0.023 & 0.549 $\pm$ 0.001 & 0.757 $\pm$ 0.035 & \textbf{0.837} $\pm$ 0.127 & 0.600 $\pm$ 0.100 & 0.775 $\pm$ 0.124 \\
            \midrule
            \multicolumn{1}{l}{Inference} & Accuracy (\%) & \textbf{0.849} $\pm$ 0.004 & 0.692 $\pm$ 0.003 & 0.819 $\pm$ 0.044 & \textbf{0.911} $\pm$ 0.068 & 0.744 $\pm$ 0.174 & 0.789 $\pm$ 0.146 \\
            & F1 score & \textbf{0.831} $\pm$ 0.005 & 0.544 $\pm$ 0.002 & 0.808 $\pm$ 0.039 & \textbf{0.883} $\pm$ 0.103 & 0.680 $\pm$ 0.244 & 0.761 $\pm$ 0.113 \\
            \bottomrule
        \end{tabular}%
    }
    \label{tab:simulation}%
\end{table*}%

\paragraph{Lorenz attractor}
Lorenz attractor is a canonical nonlinear dynamical system with the following nonlinear dynamic for $z_t = [z_{t,1},z_{t,2},z_{t,3}]$:
\begin{equation}
    \frac{d \mathbf{z}}{d t}=\left[\begin{array}{c}
        \alpha\left(\mathrm{z}_{2}-\mathrm{z}_{1}\right) \\
        \mathrm{z}_{1}\left(\beta-\mathrm{z}_{1}\right)-\mathrm{z}_{2} \\
        \mathrm{z}_{1} \mathrm{z}_{2}-\gamma \mathrm{z}_{3}
    \end{array}\right]
\end{equation}
The variable $z_t=[z_{t,1},z_{t,2},z_{t,3}]^T$ is treated as a latent variable and thus is unobservable. In the simulation, we considered a 10-dimensional time series  $y_t = Wz_t + v_t$, where $W \in R^{10\times3}$, $v_{t} \sim N(0,0.5I_{10})$. The same dataset was used in \cite{farnoosh2020deep}. Similar to the toy example, we set $x_t = y_{t-1}$. The traces of the Lorenz attractor can be roughly separated into two ellipses. We simulated a time series with a length of 3000 and transformed the time series into subsequences with a length of 5, resulting in 2990 subsequences. The first 1000, the following 990, and the last 1000 subsequences are used for training, validation, and testing respectively. 

The forecasted switching variables of the DS$^3$M are shown in Figure \ref{fig:lorenzswitching}. The model successfully separates the two ellipses with a forecasting accuracy of 0.882  $\pm$  0.079 (a relative improvement of 43.14\% and 11.99\% compared with SNLDS and DSARF respectively) and an F1 score of 0.837  $\pm$  0.127 (a relative improvement of 39.50\% and 8.11\%), see Table \ref{tab:simulation}.  For the forecasting accuracy of the observations, the DS$^3$M has smaller RMSE and MAPE compared to SNLDS, but did not beat DSARF. The superior performance of the DSARF is because the simulated dataset is generated as a multiplication of weights and factors, which fits the assumption of the generative model of the DSARF.  As for the segmentation task (inference),  the DS$^3$M also achieves the highest accuracy: 0.911  $\pm$  0.068 (a relative improvement of 22.51\% and 15.49\%) and best F1 score: 0.883  $\pm$  0.103 (a relative improvement of 29.97\% and 16.06\%).

We conducted additional experiments by setting the number of switching states to 3 for this simulated dataset. The results show that it succeeds in learning to only use two states and ignore the redundant state. The redundant state has a very small average predictive probability (0.073) over the test samples, while for the other two states, the average predictive probability is 0.512 and 0.414 respectively. 

\subsection{Real  data analysis}

We conducted a thorough evaluation of DS$^3$M's performance across six real-world datasets, encompassing diverse fields. These datasets are: \textbf{Sleep} Apnea,  the US \textbf{unemployment} rate,  \textbf{Hangzhou} metro, \textbf{Seattle} traffic and \textbf{Pacific} surface temperature, and French \textbf{electricity} demand. These datasets not only span a range of disciplines but also exhibit varying data characteristics in terms of sampling frequency and dimensionality. Here is a brief overview of each dataset:
\begin{itemize}
    \item[-] the \textbf{Sleep} Apnea dataset is a public physiological dataset from a patient diagnosed with sleep apnea, a medical condition in which patients intermittently stop breathing during sleep. The respiration pattern in sleep apnea can be characterized by at least two regimes --  no breathing and gasping breathing induced by reflex arousal.  We use the same separation of training and testing data as in \cite{ghahramani2000variational} and \cite{farnoosh2020deep}. 	 
    \item[-]The monthly US \textbf{Unemployment} rate\footnote{US Unemployment Rate,  https://fred.stlouisfed.org/series/UNRATE}   is one of the most important indicators of the US economy. The data are from January 1948  to March 2021  and the last 20 years are used for testing. 
    \item[-]The \textbf{Hangzhou} Metro dataset \footnote{Hangzhou Incoming Passenger Flow, https://tianchi.aliyun. com/competition/entrance/231708/} consists of the incoming passenger flow of 80 metro stations in Hangzhou, China from January 1 to January 25, 2019 \citep{farnoosh2020deep}.  The passenger flow data have a temporal resolution of 10-minutes during the service hour, i.e. 108 points per day. The last 5 days are used for testing. 
    \item[-]The \textbf{Seattle} Traffic dataset\footnote{Seattle Inductive Loop Detector Dataset, https://github.com/zhiyongc/Seattle-Loop-Data} contains the traffic speed from 323 loop detectors in Seattle, USA, from January 1 to January 28, 2015 \citep{farnoosh2020deep}. It has a temporal resolution of 5-min, i.e. 288 points in a day. The last 5 days are reserved for testing. 
    \item[-]The \textbf{Pacific} surface temperature dataset\footnote{Pacific Ocean Temperature Dataset, http://iridl.ldeo.columbia.edu/} consists of monthly surface temperatures of the Pacific for 2520 gridded spatial locations from January 1970 to December 2002 \citep{farnoosh2020deep}. The last 5 years are used for testing.
    \item[-]The French \textbf{Electricity} demand dataset contains half-hourly electricity demand in France from January 1, 2012  to December 31, 2019, which is also used in \cite{xu2020probabilistic,Cho2013}. The year 2019 is used for testing. For this dataset, we only have one time series, and the testing data spans one year. 
\end{itemize}	
A summary of the data sets is provided in Table \ref{tab:dataset} and more details are given in the supplementary materials. We performed both short-term and long-term forecasting. For short-term prediction, we conducted one-step ahead forecasting with rolling windows. All models were trained for 100 epochs on the data preceding the test data and remained fixed during the forecasting of the test data. For long-term prediction, we performed forecasting for the entire test data sequentially, starting from the beginning of the test dataset.

\begin{table}[H]
    \centering
    \caption{Description of the datasets}
    \resizebox{0.5\textwidth}{!}{ 
        \begin{tabular}{clrrl}
            \toprule
            \multicolumn{1}{l}{\textbf{Dataset}} & \textbf{frequency} & \multicolumn{1}{l}{\textbf{$D$}} & \multicolumn{1}{l}{$T$+$T$\_test} & \multicolumn{1}{l}{{$T$\_test}} \\
            \midrule
            \textbf{Sleep} & half a second & 1     & 2000      & 1000 (500 seconds) \\
            \textbf{Unemployment} & month & 1     &   879    & \multicolumn{1}{l}{240 (20 years)} \\
            \textbf{Hangzhou} &  10 mins & 80    & 2700  & \multicolumn{1}{l}{540 (5 days)} \\
            \textbf{Seattle} &  5 mins & 323   & 8064  & \multicolumn{1}{l}{1440 (5 days)} \\
            \textbf{Pacific} & month & 2520  & 396   & \multicolumn{1}{l}{60 (5 years)} \\
            \textbf{Electricity} & half a hour & 48    &2921    & 320  days (1 year)\\
            \bottomrule
        \end{tabular}%
    }
    \label{tab:dataset}%
\end{table}%

\paragraph{Short-term prediction results}
\begin{table*}[htbp]
    \centering
    \caption{Comparison of RMSE and MAPE on testing data. The best models are in bold. ``-" indicates the model forecasts diverge to unreasonable values and are omitted.} 
    \resizebox{\textwidth}{!}{  
        \begin{tabular}{ccrrrrrrrrrr}
            \toprule
            &       & \multicolumn{5}{c}{\textbf{RMSE}}     & \multicolumn{5}{c}{\textbf{MAPE (\%)}} \\
            \cmidrule(lr){3-7} \cmidrule(lr){8-12}          & \textbf{Datasets} & \multicolumn{1}{c}{\textbf{DS$^3$M}} & \multicolumn{1}{c}{\textbf{SNLDS}} & \multicolumn{1}{c}{\textbf{DSARF}} & \multicolumn{1}{c}{\textbf{SRNN}} & \multicolumn{1}{c}{\textbf{GRU}} & \multicolumn{1}{c}{\textbf{DS$^3$M}} & \multicolumn{1}{c}{\textbf{SNLDS}} & \multicolumn{1}{c}{\textbf{DSARF}} & \multicolumn{1}{c}{\textbf{SRNN}} & \multicolumn{1}{c}{\textbf{GRU}} \\
            \cmidrule(lr){1-7} \cmidrule(lr){8-12}    
            \multirow{6}[2]{*}{\textbf{Short-term}} & \textbf{Sleep} & \textbf{1201} & 2789  & 1557  & 1806  & \underline{1264}  & \textbf{15.46} & 88.06 & 39.25 & 50.8  & \underline{31.17} \\
            & \textbf{Unemployment} & \textbf{0.75} & 1.59  & 1.06  & 2.01  & \underline{1.05}  & \textbf{4.53} & 16.13 & 8.11  & 23.15 & \underline{5.13} \\
            & \textbf{Hangzhou} & \textbf{32.53} & 36.67 & 34.81 & \underline{33.80} & 38.42 & \underline{24.04} & \textbf{23.90} & 29.73 & 25.40 & 30.48 \\
            & \textbf{Seattle} & \textbf{4.16} & 4.18  & 4.44  & \underline{4.17}  & 4.18  & \textbf{5.81} & \underline{5.85}  & 7.27  & 6.00  & 6.89 \\
            & \textbf{Pacific} & 0.57  & 15.78 & \textbf{0.53} & 0.58  & \underline{0.56}  & 1.69  & 58.01 & \textbf{1.57} & 1.74  & \underline{1.68} \\
            & \textbf{Electricity} & \textbf{2971} & 5133  & 8805  & \underline{3642}  & 4784  & \textbf{4.58} & 7.79  & 18.64 & \underline{5.34}  & 6.60 \\
            \cmidrule(lr){1-7} \cmidrule(lr){8-12}   
            \multirow{4}[2]{*}{\textbf{Long-term}} & \textbf{Hangzhou} & 47.50 & \underline{42.83} & \textbf{42.28} & 60.89 & 73.18 & \textbf{38.20} & 50.6  & \underline{43.65} & 82.81 & 86.61 \\
            & \textbf{Seattle} & \textbf{4.17} & 4.19  & -     & \textbf{4.17}  & 16.93 & \textbf{5.81} & 5.86  & -     & \textbf{5.81}  & 27.95 \\
            & \textbf{Pacific} & \textbf{0.72} & -     & \underline{0.73}  & 0.98  & 0.76 & \textbf{2.15} & -     & {2.29}  & 2.99  & 2.22 \\
            
            \bottomrule
        \end{tabular}%
    }
    \label{tab:results}%
\end{table*}%

Table \ref{tab:results} presents the prediction results for the six datasets in terms of both MAPE (Mean Absolute Percentage Error) and RMSE (Root Mean Square Error), as defined in the supplementary materials. DS$^3$M demonstrates superior performance across the board. Specifically, for the Sleep, Unemployment rate, Seattle, and Electricity datasets, DS$^3$M outperforms all alternative models in terms of both RMSE and MAPE. Notably, the RMSE values exhibit relative improvements ranging from 5.0\%-56.9\%, 28.9\%-62.8\%, 0.5\%-6.3\%, and 18.4\%-66.3\% for these four datasets, while the MAPE values demonstrate reductions of 15.71\%-72.60\%, 0.60\%-18.62\%, 0.09\%-1.46\%, and 0.76\%-14.06\% against the alternative models. For the Hangzhou dataset, DS$^3$M achieves the lowest RMSE and exhibits comparable MAPE to SNLDS (the best-performing model for this dataset). For the Pacific dataset, DS$^3$M attains competitive performance comparable to DSARF, which holds the best performance in this context. We also conducted the DM test \citep{diebold2002comparing} to check whether the forecast errors from the models are significantly different. The test confirms that the DS$^3$M model significantly outperforms the alternatives for the Sleep, Unemployment rate, and Electricity datasets according to the DM test. There is no significant difference between the DS$^3$M and the best alternatives for the Seattle, Hangzhou, and Pacific datasets. 

\begin{figure*}
    \centering
    \small
    \begin{subfigure}{0.8\textwidth}
        \includegraphics[width=1\linewidth]{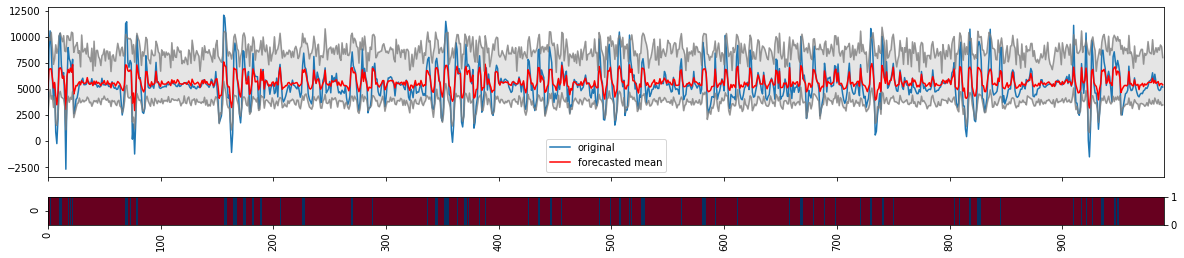}
        \caption{Sleep apnea (measured at 2 Hz)}
        \label{fig:apnea}
    \end{subfigure}
    \begin{subfigure}{0.8\textwidth}
        \includegraphics[width=1\linewidth]{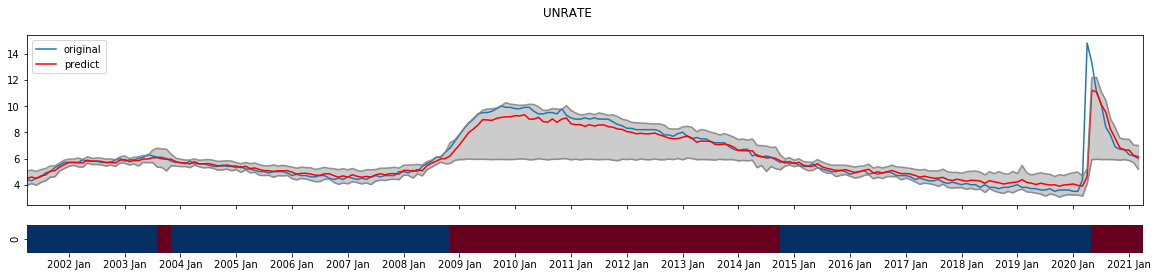}
        \caption{US unemployment rate}
        \label{fig:unrate}
    \end{subfigure}
    \begin{subfigure}{0.4\textwidth}
        \centering
        \includegraphics[width=\linewidth]{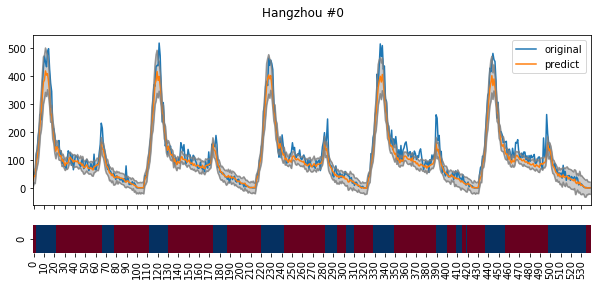}
        \caption{ Hangzhou metro station 0}
        \label{fig:hangzhou1}
    \end{subfigure}
    \begin{subfigure}{0.4\textwidth}
        \centering
        \includegraphics[width=\linewidth]{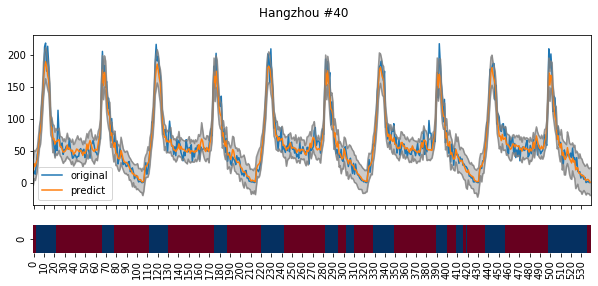}
        \caption{ Hangzhou metro station 40}
        \label{fig:hangzhou2}
    \end{subfigure}
    
    \begin{subfigure}[b]{0.4\textwidth}
        \centering
        \includegraphics[width=\textwidth]{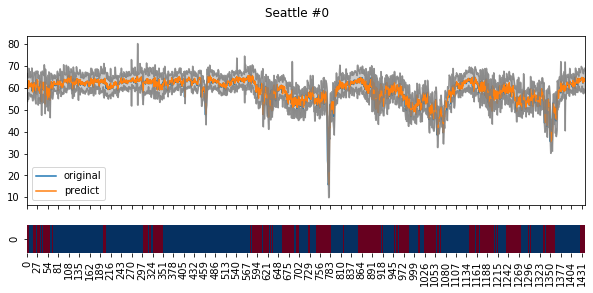}
        \caption{ Seattle traffic loop 0}
        \label{fig:seattle1}
    \end{subfigure}
    \begin{subfigure}[b]{0.4\textwidth}
        \centering
        \includegraphics[width=\textwidth]{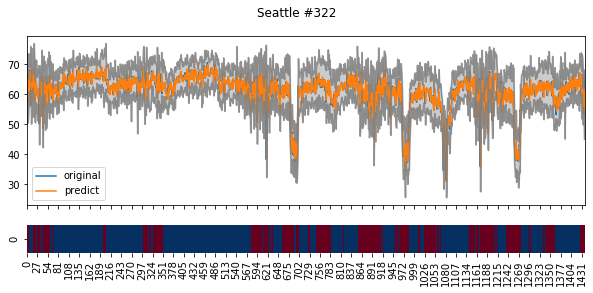}
        \caption{ Seattle traffic loop 322}
        \label{fig:seattle2}
    \end{subfigure}
    
    \begin{subfigure}[b]{0.4\textwidth}
        \centering
        \includegraphics[width=\textwidth]{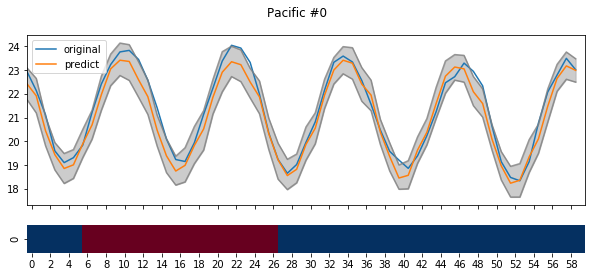}
        \caption{ Pacific location  0}
        \label{fig:pacific1}
    \end{subfigure}
    \begin{subfigure}[b]{0.4\textwidth}
        \centering
        \includegraphics[width=\textwidth]{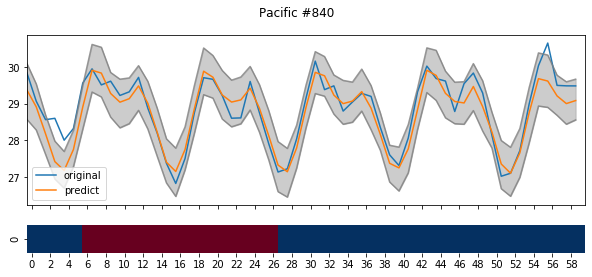}
        \caption{  Pacific location  840}
        \label{fig:pacific2}
    \end{subfigure}
    
    \begin{subfigure}[b]{0.4\textwidth}
        \centering
        \includegraphics[width=\textwidth]{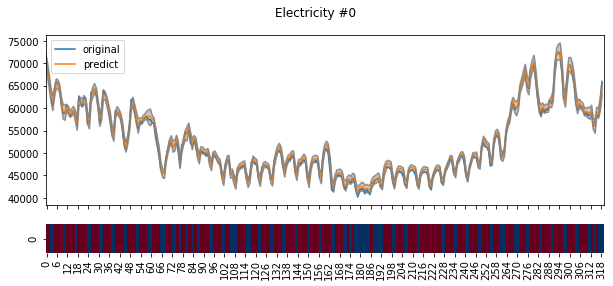}
        \caption{French Electricity  0:00}
        \label{fig:electricity1}
    \end{subfigure}
    \begin{subfigure}[b]{0.4\textwidth}
        \centering
        \includegraphics[width=\textwidth]{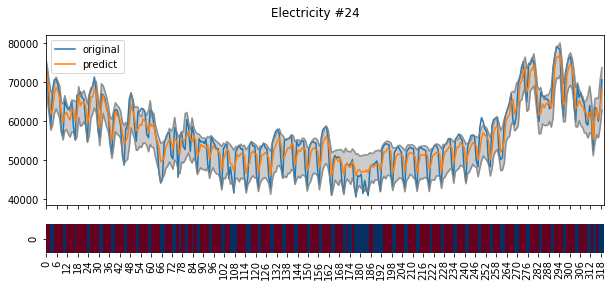}
        \caption{French Electricity 12:00}
        \label{fig:electricity2}
    \end{subfigure}
    \caption{Predictions of the testing data for different datasets}
    \label{fig:results}
\end{figure*}

Figure \ref{fig:results} visually presents the short-term predictions of the testing data along with the identified switching regimes for various datasets. For the \textbf{Sleep} dataset, DS$^3$M segregates the time series into two distinct regimes represented by blue and red shades. Notably, the consistent red regime is found to correspond to periods when the patient experiences little to no breathing, while the blue regime corresponds to periods marked by gasping breaths. In the case of \textbf{Unemployment} rates, DS$^3$M successfully separates the time series into two regimes. The red regimes align with times of elevated unemployment, notably during the 2009 financial crisis and the 2020 Covid-19 pandemic. Further, specific illustrations showcase values at randomly selected locations for datasets such as Hangzhou, Seattle, Pacific, and French Electricity demand. DS$^3$M's automated segmentation of Hangzhou traffic data into peak and non-peak hours is particularly noteworthy. In the context of the \textbf{Seattle} dataset, the red regime signifies periods of heightened traffic volatility. In the \textbf{Pacific} dataset, different regimes are associated with shifts in the time series level. For instance, at Location 0, the red regime exhibits a higher level compared to the blue regime, while the reverse is true for Location 840. Lastly, for the French \textbf{Electricity} demand dataset, the red regime is found to align predominantly with working days, while the blue regime corresponds to weekends. In general, it shows that all the predicted values trace the true values closely and the 90\% confidence intervals cover most of the true values in the future.  
It is worth noting that unlike models that involve multiple features and use metrics such as Shapley values for feature importance {\color{blue} \citep{babaei2025rank}}, DS$^3$M focuses on interpreting latent regimes only using lagged values of the time series. Its main contribution lies in the ability to precisely identify latent regimes, providing insights into changes in dynamic behavior. This interpretability centers on understanding the dynamics and changes in latent regimes over time. By modeling the switching behavior and capturing transitions between different regimes, DS$^3$M is able to provide insights into the underlying processes driving the data and provide economic interpretation, in addition to the enhanced forecast accuracy.

\paragraph{Long-term prediction results} 
In the long-term prediction experiment, we excluded the Sleep and Unemployment datasets due to their lack of periodic patterns and high chaotic nature, making them unsuitable for long-term prediction assessment. Similarly, the Electricity dataset was omitted from the long-term prediction analysis, given its test length spanning one year, which renders long-term predictions less practical for high-frequency data across all models.

Table \ref{tab:results} showcases the long-term forecasting errors for the various datasets. Notably, DS$^3$M exhibits superior performance in terms of RMSE for the Pacific dataset. For both Hangzhou and Pacific datasets, DS$^3$M outperforms the alternatives in terms of MAPE. Regarding the Seattle dataset, DS$^3$M and SRNN demonstrate comparable performance. These findings highlight DS$^3$M's favorable outcomes in the context of long-term predictions, especially for the specific datasets mentioned. 
The DM test shows that the DS$^3$M model significantly outperforms the best alternatives for the Pacific dataset, while there is no significant difference for the Seattle dataset. For the Hangzhou dataset, DSARF slightly outperforms DS$^3$M, but in terms of MAPE, DS$^3$M performs better. 

 These results indicate that although DS$^3$M is not consistently the best model, it is often the best or second-best model. The performance of DS$^3$M is generally more robust compared to the alternatives.

\section{Conclusion} \label{sec:conclusion}
We proposed the deep switching state space model (DS$^3$M) for forecasting nonlinear time series characterized by regime switching. DS$^3$M effectively captures these intricate dynamics by utilizing both discrete and continuous latent variables in conjunction with recurrent neural networks. This distinctive approach combines the power of deep learning with stochastic latent variable models, enabling accurate and interpretable forecasting.

A key strength of DS$^3$M lies in its versatility across diverse datasets. The model's architecture, comprised of a recurrent neural network (RNN) and a nonlinear switching state space model (SSSM), is capable of accommodating small and large data alike. The amortized variational inference method, employed for estimation, trains both the inference and generative networks together, ensuring applicability across varying data scales. The DS$^3$M's efficacy is demonstrated across a range of simulated and real-world datasets, showcasing its competitive performance relative to several state-of-the-art methods.

There are some limitations of the proposed model. Firstly, an open loop of the transition of the discrete variable is not considered, as we found during the experiments that the current open-loop design by making the discrete latent variables always depend on the continuous latent variables and/or observations may lead to unnecessarily frequent switching of the latent variables. If there is a time series with ultra-frequent regime switching behaviors, such a recurrent structure may be useful.  More sophisticated architecture can be designed to account for this and we leave it for future research. Secondly, it is challenging to choose the number of switching states. This is an open question in the literature. Future work can use a Dirichlet prior to automatically deciding the number of switching states.

These contributions establish DS$^3$M as a robust and adaptable tool for forecasting complex nonlinear time series, offering a bridge between the worlds of deep learning and latent variable models. As this field evolves, DS$^3$M is positioned to enhance our ability to model and understand intricate dynamics in diverse applications.

\section*{Acknowledgments}
We acknowledge the financial support provided by grants A-8000828-00-00 and A-8000828-01-00 (Regime-Switching Markov Decision Process with Applications in Digital FinTech), as well as grant A-8000014-00-00 (Deep State Space Models for Non-stationary Time Series).

\bibliographystyle{elsarticle-num-names}
\bibliography{refxx}

\appendix
\section{Derivation of the evidence lower bound (\mbox{ELBO})} \label{appen: elbo}
    The \mbox{ELBO} for the log-likelihood can be derived as follows:
    \begin{equation*}\tiny
        \resizebox{1.2\linewidth}{!}{$
            \begin{aligned}
                \mathcal{L}(\theta) 
                &\geq \mbox{ELBO}(\theta,\phi)\\	
                &= \iint q_{\phi}\left(\mathbf{z}_{1 : T}, \mathbf{d}_{1 : T} | \mathbf{y}_{1 : T}, \mathbf{x}_{1 : T}\right) \log \frac{p_{\theta}\left(\mathbf{y}_{1 : T}, \mathbf{z}_{1 : T}, \mathbf{d}_{1 : T} | \mathbf{x}_{1 : T}\right)}{q_{\phi}\left(\mathbf{z}_{1 : T}, \mathbf{d}_{1 : T} | \mathbf{y}_{1 : T}, \mathbf{x}_{1 : T}\right)} \mathrm{d} \mathbf{z}_{1 : T} \mathrm{d} \mathbf{d}_{1 : T} \\
                &= \iint q_{\phi}\left(\mathbf{z}_{1 : T}, \mathbf{d}_{1 : T} | \mathbf{y}_{1 : T}, \mathbf{h}_{1 : T}\right) \log \frac{p_{\theta}\left(\mathbf{y}_{1 : T}, \mathbf{z}_{1 : T}, \mathbf{d}_{1 : T} | \mathbf{h}_{1 : T}\right)}{q_{\phi}\left(\mathbf{z}_{1 : T}, \mathbf{d}_{1 : T} | \mathbf{y}_{1 : T}, \mathbf{h}_{1 : T}\right)} \mathrm{d} \mathbf{z}_{1 : T} \mathrm{d} \mathbf{d}_{1 : T} \\
                &=\mathbb{E}_{q_{\phi}}\left[\log p_{\theta}\left(\mathbf{y}_{1 : T} | \mathbf{z}_{1 : T},\mathbf{d}_{1 : T}, \mathbf{h}_{1 : T}\right)\right] -\mathrm{KL}\left(q_{\phi}\left(\mathbf{z}_{1 : T},\mathbf{d}_{1 : T}| \mathbf{y}_{1 : T},\mathbf{h}_{1 : T}\right) \| p_{\theta}\left(\mathbf{z}_{1 : T}, \mathbf{d}_{1 : T}| \mathbf{h}_{1 : T} \right)\right)\\
                &=\mathbb{E}_{q_{\phi}}\left[\sum_{t}\log  p_{\theta}\left(\mathbf{y}_{t} | \mathbf{z}_{t},\mathbf{d}_{t}, \mathbf{h}_{t}\right)\right]  -  \int q_{\phi}\left(\mathbf{z}_{1 : T},\mathbf{d}_{1 : T}| \mathbf{y}_{1 : T},\mathbf{h}_{1 : T}\right) \log{\frac{q_{\phi}\left(\mathbf{z}_{1 : T},\mathbf{d}_{1 : T}| \mathbf{y}_{1 : T},\mathbf{h}_{1 : T}\right)}{ p_{\theta}\left(\mathbf{z}_{1 : T}, \mathbf{d}_{1 : T}| \mathbf{h}_{1 : T} \right)}}\mathrm{d} \mathbf{z}_{1 : T} \mathrm{d} \mathbf{d}_{1 : T}\\
                &=\sum_{t}\mathbb{E}_{q_{\phi}}\left[\log  p_{\theta}\left(\mathbf{y}_{t} | \mathbf{z}_{t}, \mathbf{d}_{t}, \mathbf{h}_{t}\right)\right] \\
                &\quad \quad -  \sum_{t}\int q_{\phi}\left(\mathbf{z}_{1 : T},\mathbf{d}_{1 : T}| \mathbf{y}_{1 : T},\mathbf{h}_{1 : T}\right)  \log{\frac{ q_{\phi_z}\left(\mathbf{z}_{t} | \mathbf{z}_{t-1}, \mathbf{d}_{t}, \mathbf{A}_t\right)q	_{\phi_d}\left(\mathbf{d}_{t} | \mathbf{d}_{t-1}, \mathbf{A}_t\right)}{ p_{\theta}\left(\mathbf{z}_{t}|\mathbf{z}_{t-1},\mathbf{d}_{t}, \mathbf{h}_{t} \right)p_{\theta}(\mathbf{d}_{t}|\mathbf{d}_{t-1})}}\mathrm{d} \mathbf{z}_{1 : T} \mathrm{d} \mathbf{d}_{1 : T}\\
                &=\sum_{t}\mathbb{E}_{q_{\phi}}\left[\log  p_{\theta}\left(\mathbf{y}_{t} | \mathbf{z}_{t}, \mathbf{d}_{t}, \mathbf{h}_{t}\right)\right] \\
                &\quad \quad \quad-  \sum_{t}\int q_{\phi}\left(\mathbf{z}_{1 : T},\mathbf{d}_{1 : T}| \mathbf{y}_{1 : T},\mathbf{h}_{1 : T}\right)  \log{\frac{ q_{\phi_z}\left(\mathbf{z}_{t} | \mathbf{z}_{t-1}, \mathbf{d}_{t}, \mathbf{A}_t\right)}{p_{\theta}\left(\mathbf{z}_{t}|\mathbf{z}_{t-1},\mathbf{d}_{t}, \mathbf{h}_{t} \right)}}\mathrm{d} \mathbf{z}_{1 : T} \mathrm{d} \mathbf{d}_{1 : T}\\
                &\quad \quad \quad-  \sum_{t}\int q_{\phi}\left(\mathbf{z}_{1 : T},\mathbf{d}_{1 : T}| \mathbf{y}_{1 : T},\mathbf{h}_{1 : T}\right)  \log{\frac{ q	_{\phi_d}\left(\mathbf{d}_{t} | \mathbf{d}_{t-1}, \mathbf{A}_t\right)}{ p_{\theta}(\mathbf{d}_{t}|\mathbf{d}_{t-1})}}\mathrm{d} \mathbf{z}_{1 : T} \mathrm{d} \mathbf{d}_{1 : T}\\
                &=\sum_{t}\mathbb{E}_{q_{\phi}^*(\mathbf{z}_{t},\mathbf{d}_{t})}\left[\log  p_{\theta}\left(\mathbf{y}_{t} | \mathbf{z}_{t}, \mathbf{d}_{t}, \mathbf{h}_{t}\right)\right] \\
                &\quad \quad-  \sum_{t} \mathbb{E}_{q_{\phi}^*(\mathbf{z}_{t-1},\mathbf{d}_{t-1})}\int q_{\phi_z}\left(\mathbf{z}_{t}| \mathbf{z}_{t-1}, \mathbf{d}_{t}, \mathbf{A}_t\right)q_{\phi_d}\left(\mathbf{d}_{t}  | \mathbf{d}_{t-1}, \mathbf{A}_t\right) \log{\frac{ q_{\phi_z}\left(\mathbf{z}_{t} | \mathbf{z}_{t-1}, \mathbf{d}_{t}, \mathbf{A}_t\right)}{ p_{\theta}\left(\mathbf{z}_{t}|\mathbf{z}_{t-1},\mathbf{d}_{t}, \mathbf{h}_{t} \right)}}\mathrm{d} \mathbf{z}_{t} \mathrm{d} \mathbf{d}_{t}\\
                &\quad \quad-  \sum_{t} \mathbb{E}_{q_{\phi}^*(\mathbf{d}_{t-2})}\int q_{\phi_d}\left(\mathbf{d}_{t-1}  | \mathbf{d}_{t-2}, A_{t-1}\right) q_{\phi_d}\left(\mathbf{d}_{t}  | \mathbf{d}_{t-1}, \mathbf{A}_t\right) \log{\frac{ q	_{\phi_d}\left(\mathbf{d}_{t} | \mathbf{d}_{t-1}, \mathbf{A}_t\right)}{ p_{\theta}(\mathbf{d}_{t}|\mathbf{d}_{t-1})}}\mathrm{d} \mathbf{d}_{t}\\
                &=\sum_{t}\mathbb{E}_{q_{\phi}^*(\mathbf{z}_{t-1},\mathbf{d}_{t-1})}\mathbb{E}_{q_{\phi_z}(\mathbf{z}_{t}|\mathbf{z}_{t-1},\mathbf{d}_{t},\mathbf{A}_t)}\mathbb{E}_{q_{\phi_d}(\mathbf{d}_{t}|\mathbf{d}_{t-1},\mathbf{A}_t)}\left[\log  p_{\theta}\left(\mathbf{y}_{t} | \mathbf{z}_{t},\mathbf{d}_{t}, \mathbf{h}_{t}\right)\right] \\
                &\quad \quad-  \sum_{t} \mathbb{E}_{q_{\phi}^*(\mathbf{z}_{t-1},\mathbf{d}_{t-1})}\mathbb{E}_{q_{\phi_d}\left(\mathbf{d}_{t} | \mathbf{d}_{t-1}, \mathbf{A}_t\right)}KL\left[ q_{\phi_z}\left(\mathbf{z}_{t} | \mathbf{z}_{t-1}, \mathbf{d}_{t}, \mathbf{A}_t\right)\| p_{\theta}\left(\mathbf{z}_{t}|\mathbf{z}_{t-1},\mathbf{d}_{t}, \mathbf{h}_{t} \right)\right] \\
                &	\quad \quad \quad-  \sum_{t} \mathbb{E}_{q_{\phi}^*(\mathbf{d}_{t-2})}\mathbb{E}_{ q_{\phi_d}}\left(\mathbf{d}_{t-1}  | \mathbf{d}_{t-2}, A_{t-1}\right)  KL\left[ q	_{\phi_d}\left(\mathbf{d}_{t} | \mathbf{d}_{t-1}, \mathbf{A}_t\right)\| p_{\theta}(\mathbf{d}_{t}|\mathbf{d}_{t-1})\right]\\
                &=\sum_{t}\Biggl\{\mathbb{E}_{q_{\phi}^*(\mathbf{z}_{t-1},\mathbf{d}_{t-1})}\sum_{\mathbf{d}_t}{q_{\phi_d}(\mathbf{d}_{t}|\mathbf{d}_{t-1},\mathbf{A}_t)}\mathbb{E}_{q_{\phi_z}(\mathbf{z}_{t}|\mathbf{z}_{t-1},\mathbf{d}_{t},\mathbf{A}_t)}\left[\log  p_{\theta}\left(\mathbf{y}_{t} | \mathbf{z}_{t}, \mathbf{d}_{t},\mathbf{h}_{t}\right)\right] \\
                &\quad \quad-\mathbb{E}_{q_{\phi}^*(\mathbf{z}_{t-1},\mathbf{d}_{t-1})} \sum_{\mathbf{d}_t}{q_{\phi_d}\left(\mathbf{d}_{t} | \mathbf{d}_{t-1}, \mathbf{A}_t\right)}KL\left[ q_{\phi_z}\left(\mathbf{z}_{t} | \mathbf{z}_{t-1}, \mathbf{d}_{t}, \mathbf{A}_t\right)\| p_{\theta}\left(\mathbf{z}_{t}|\mathbf{z}_{t-1},\mathbf{d}_{t}, \mathbf{h}_{t} \right)\right] \\
                &\quad \quad \quad- \mathbb{E}_{q_{\phi}^*(\mathbf{d}_{t-2})} \sum_{\mathbf{d}_{t-1}}{q_{\phi_d}\left(\mathbf{d}_{t-1} | \mathbf{d}_{t-2}, A_{t-1}\right)}KL\left[ q	_{\phi_d}\left(\mathbf{d}_{t} | \mathbf{d}_{t-1}, \mathbf{A}_t\right)\| p_{\theta}(\mathbf{d}_{t}|\mathbf{d}_{t-1})\right]\Biggr\} 
            \end{aligned}
            $}
    \end{equation*}

\section{Stability analysis of DS$^3$M}
\noindent{\bf Theorem 1}\\
DS\textsuperscript{3}M exhibits latent state dynamics \( z_t \sim N(\mu_{z,t}^{(d_t)}, \Sigma_{z,t}^{(d_t)}) \), where the mean \( \mu_{z,t}^{(d_t)} \) and the diagonal covariance matrix \( \Sigma_{z,t}^{(d_t)} \) are parameterized by two neural network models \( f^{(d_t)}_{z,1} \) and \( f^{(d_t)}_{z,2} \). The latent variable \( z_t \) is globally stable in mean-square under the following conditions:
\begin{flalign} 
& ||W_i^{f^{(d_t)}_{z,1}}||_2 < 1, \quad \textrm{for} \ 1 \leq i \leq L_{f^{(d_t)}_{z,1}}, \label{w_constrain_1}\\
& ||\mathcal{A}_i^{f^{(d_t)}_{z,1}}||_2 \leq 1, \quad \textrm{for} \ 1 \leq i \leq L_{f^{(d_t)}_{z,1}}-1, \label{acti_constrain_1}\\
& ||W_j^{f^{(d_t)}_{z,2}}||_2 < 1, \quad \textrm{for} \ 1 \leq j \leq L_{f^{(d_t)}_{z,2}}, \label{w_constrain_2}\\
& ||\mathcal{A}_j^{f^{(d_t)}_{z,2}}||_2 \leq 1, \quad \textrm{for} \ 1 \leq j \leq L_{f^{(d_t)}_{z,2}}-1. \label{acti_constrain_2}
\end{flalign}
Similarly, for the observed time series, we model $y_t \sim \mathcal{N}(\mu_{y,t}^{(d_t)}, \Sigma_{y,t}^{(d_t)})$, where the mean $\mu_{y,t}^{(d_t)}$ and the covariance matrix $\Sigma_{y,t}^{(d_t)}$ are parameterized by neural networks $f^{(d_t)}_{y,1}$ and $f^{(d_t)}_{y,2}$, respectively. The observed variable \( y_t \) is globally stable in mean-square under the following conditions:
\begin{flalign} 
& ||W_i^{f^{(d_t)}_{y,1}}||_2 < 1, \quad \textrm{for} \ 1 \leq i \leq L_{f^{(d_t)}_{y,1}}, \label{w_constrain_3}\\
& ||\mathcal{A}_i^{f^{(d_t)}_{y,1}}||_2 \leq 1, \quad \textrm{for} \ 1 \leq i \leq L_{f^{(d_t)}_{y,1}}-1, \label{acti_constrain_3}\\
& ||W_j^{f^{(d_t)}_{y,2}}||_2 < 1, \quad \textrm{for} \ 1 \leq j \leq L_{f^{(d_t)}_{y,2}}, \label{w_constrain_4}\\
& ||\mathcal{A}_j^{f^{(d_t)}_{y,2}}||_2 \leq 1, \quad \textrm{for} \ 1 \leq j \leq L_{f^{(d_t)}_{y,2}}-1. \label{acti_constrain_4}
\end{flalign}
Here, \( W_i^{f^{(d_t)}_{z,1}}, W_j^{f^{(d_t)}_{z,2}}, W_i^{f^{(d_t)}_{y,1}} \) and \( W_j^{f^{(d_t)}_{y,2}} \) represent the weight matrices of neural networks \( f^{(d_t)}_{z,1}, f^{(d_t)}_{z,2}, f^{(d_t)}_{y,1} \) and \( f^{(d_t)}_{y,2} \), respectively; meanwhile, \(\mathcal{A}_i^{f^{(d_t)}_{z,1}}, \mathcal{A}_j^{f^{(d_t)}_{z,2}}, \mathcal{A}_i^{f^{(d_t)}_{y,1}} \) and \( \mathcal{A}_j^{f^{(d_t)}_{y,2}} \) denote their corresponding activation scaling matrices.

\noindent{$Proof$}:\\
\cite{nandanoori2018mean} and \cite{drgona2021stochastic} describes that a dynamic system is mean-square stable if its expected value and covariance converge over time. That is, a time series $\mathbf{a_t} \in \mathbb{R}^n $ is stable if $\mathbf{a_t}$ has $\lim_{t \to \infty} \mathbb{E}(\mathbf{a_t}) = \mu$ and $\lim_{t \to \infty} \mathbb{E}(\mathbf{cov(a_t,a_t)}) = \mathbf{\Sigma}$, in which $\mu \in \mathbb{R}$ and $\mathbf{\Sigma} \in \mathbb{R}^n$. 

Denote a neural network $f: \mathbb{R}^{n} \rightarrow \mathbb{R}^{m}$ parameterized by $\theta = \{W_0,...,W_L,b_0,...,b_L\}$ with hidden layers $1 \leq l \leq L$ with bias given as follows:
\begin{align} 
f_{\theta}(x) &= W_l h_l + b_l \\
h_l &= a(W_{l-1} h_{l-1} + b_{l-1})
\end{align}
with $h_0 = x$, and $a : \mathbb{R}^{k} \rightarrow \mathbb{R}^{k}$ representing element-wise application of an activation function to vector elements such that $a(z) := [a(z_1) . . . a(z_k)]$. 

Any neural network \( f_{\theta}(x) \) with arbitrary activation functions—such as \( f^{(d_t)}_{z,1},\ f^{(d_t)}_{z,2},\ f^{(d_t)}_{y,1},\ \text{and}\ f^{(d_t)}_{y,2} \), which represent the means \( \mu_{z,t}^{(d_t)} \) and \( \mu_{y,t}^{(d_t)} \) and the diagonal covariance matrices \( \Sigma_{z,t}^{(d_t)} \) and \( \Sigma_{y,t}^{(d_t)} \) of latent or observed variables—can be equivalently expressed as a pointwise affine map:
    \begin{align} 
        f_{\theta}(x) &= W_L\mathcal{A}_{L-1}(W_{L-1}(...\mathcal{A}_{1}(W_1\mathcal{A}_{0}(W_0x + b_0)\nonumber\\
        & + b_1)...) + b_{L-1}) + \mathbb{b}_L \nonumber\\
        &= W_xx + b_x 
        \label{nn_affine_map}
    \end{align}
    where $W_x$ and $b_x$ are parametrized by input vector $x$: 
    \begin{align} 
        W_x x &= W_L\mathcal{A}_{L-1}W_{L-1}...\mathcal{A}_{0}W_0x \label{weight}\\
        b_x &= W_L\mathcal{A}_{L-1}...W_2\mathcal{A}_1W_1\mathcal{A}_0b_0 \nonumber \\
        & + W_L\mathcal{A}_{L-1}...W_2\mathcal{A}_1b_1 + ... \\
        & + W_L\mathcal{A}_{L-1}b_{L-1} + b_L \nonumber
    \end{align}
    and $\mathcal{A}_l$ represents parameter varying diagonal matrix of activation patterns. 
    
We have shown that $f^{(d_t)}_{z,1}$ and $f^{(d_t)}_{z,2}$, with the latent variable $z_{t-1}$ as part of their inputs, can be expressed in the form \eqref{nn_affine_map}. Next, we prove that the latent variable $z_t$ is globally stable in the mean-square sense, provided that conditions \eqref{w_constrain_1} to \eqref{acti_constrain_2} are satisfied.

Assuming general non-square weights $W_i \in \mathbb{R}^{n_i \times m_i}$,
\begin{align}
||W_1W_2...W_m||_2 \leq ||W_1||_2\cdot||W_2||_2...||W_m||_2
\label{submultiplicativity}
\end{align}
holds because of the submultiplicativity of induced p-norms.Applying \eqref{submultiplicativity} to the linear parts \eqref{weight} of the mean neural network $f^{(d_t)}_{z,1}$ in the equivalent pointwise affine form \eqref{nn_affine_map} with \eqref{w_constrain_1} and \eqref{acti_constrain_1}, it produces $||W_i^{f^{(d_t)}_{z,1}}||_2 < 1$ over the entire domain of $f^{(d_t)}_{z,1}(x)$. Since An affine map is a contraction if the 2-norm of its linear part is bounded below one, and Banach fixed-point theorem states that every contractive map converges towards single point equilibirum, $f^{(d_t)}_{z,1}$ is a contractive neural network and provides convergent mean $\mu_{z,t}^{(d_t)}$ for $z_{t}$ as $t \rightarrow \infty$. 
    
Similarly, using \eqref{submultiplicativity} to the linear parts \eqref{weight} of the variance neural network $f^{(d_t)}_{z,2}$ in the equivalent pointwise affine form \eqref{nn_affine_map} with \eqref{w_constrain_2} and \eqref{acti_constrain_2}, it produces $||W_x^{f^{(d_t)}_{z,2}}||_2 < 1$. Hence, $f^{(d_t)}_{z,2}$ is a contractive neural network and provides convergent variance $\Sigma_{z,t}^{(d_t)}$ for $z_{t}$ as $t \rightarrow \infty$. Therefore, $z_t$ is globally mean-square stable when \eqref{w_constrain_1}, \eqref{acti_constrain_1}, \eqref{w_constrain_2} and \eqref{acti_constrain_2} are satisfied. 
    
    
The observed time series is modeled as \( y_t \sim \mathcal{N}(\mu_{y,t}^{(d_t)}, \Sigma_{y,t}^{(d_t)}) \), where the mean \( \mu_{y,t}^{(d_t)} \) and covariance matrix \( \Sigma_{y,t}^{(d_t)} \) are parameterized by neural networks \( f^{(d_t)}_{y,1} \) and \( f^{(d_t)}_{y,2} \), respectively. These neural networks take the stable latent variable \( z_t \) as part of their inputs and are equivalent to pointwise affine maps~\eqref{nn_affine_map}. Using the same method employed to prove that \( z_t \) is globally mean-square stable, we can similarly show that \( y_t \) is globally stable in the mean-square sense if \eqref{w_constrain_3}, \eqref{acti_constrain_3}, \eqref{w_constrain_4} and \eqref{acti_constrain_4} are satisfied. 

The constraints $||\mathcal{A}_i^{f^{(d_t)}_{z,1}}||_2 \leq 1 , ||\mathcal{A}_j^{f^{(d_t)}_{z,2}}||_2 \leq 1, ||\mathcal{A}_i^{f^{(d_t)}_{y,2}}||_2 \leq 1$ and $||\mathcal{A}_j^{f^{(d_t)}_{y,2}}||_2 \leq 1$ on activation scaling matrices implies Lipschitz continuous activation functions with constants $k \leq 1$. In practice, DS$^3$M satisfies these requirements by using ReLU (Rectified Linear Unit) activation functions. 

    
    
    \section{Details on the hyperparameters of experiments}
    \label{appendix:hyper}
    The DS$^3$M is implemented in PyTorch and all experiments are conducted on a V100 GPU.  As the sizes of the datasets are not large, a larger number of the hyperparameters does not lead to improvement of the performance.
    Therefore, we paid more attention to regularization methods during training instead of hyperparameter tuning. Specifically, we use Adam optimizer and the initial learning rate is set to 0.001 and is reduced by a factor of 0.1  when the validation loss has stopped improving for 10 epochs. An early stopping regularization is also implemented to stop the training when the validation loss has stopped improving for 20 epochs. All datasets are normalized before training and are transformed back for evaluation. 
    The classical linear KL annealing approach is used to increase the coefficients of the KL terms from 0.01 to 1 over the course of the training process. Since 100 epochs are sufficient for most experiments, we trained all models for 100 epochs during various evaluations.
    The number of switching states is set to 2.  The batch size is set to 64. The dimension of the continuous latent variables is set to 2 for the toy example, the sleep apnea and the unemployment rate,  3 for the Lorenz, 50 for Pacific, and 10 for other datasets. All RNN are chosen to be 1-layer GRUs with the hidden dimension being 10 for the toy example, the sleep apnea and the unemployment rate, 20 for the Lorenz, 200 for the Pacific and the dimension of the observations $D$ for other datasets.  All MLPs are 2-layers with the hidden dimension to be the same as the dimension of the outputs.    
    The hyperparameter of  SNLDS and SRNN is chosen with the same rules and regularization methods.
    For the DSARF, we follow the same hyperparameters specified in \cite{farnoosh2020deep}. 
    For the baseline simple GRU model, we use a grid search to fine-tune the parameters.  The number of GRU layers is chosen  in [1,2,3,4,5], the hidden units are chosen from $[D,2D,3D,4D,5D]$. The model with the best validation loss is used to perform forecasting.

\end{document}